\documentclass[10pt,twocolumn,letterpaper]{article}

\usepackage{iccv}
\usepackage{times}
\usepackage{epsfig}
\usepackage{graphicx}
\usepackage{amsmath}
\usepackage{amssymb}
\usepackage{bm}
\usepackage{stmaryrd}
\usepackage{multirow}
\usepackage{booktabs}
\usepackage[breaklinks=true,bookmarks=false]{hyperref}

\DeclareMathOperator*{\argmin}{argmin}
\DeclareMathOperator*{\argmax}{argmax}

\newcommand{\Rbb}{\ensuremath{\mathbb{R}}}

\newcommand{\ee}{\ensuremath{{\rm e}}}

\newcommand{\inv}[1]{\ensuremath{\frac{1}{#1}}}

\renewcommand{\leq}{\ensuremath{\leqslant}}

\newcommand{\adjoint}{\ensuremath{{\intercal}}}
\newcommand{\ma}[1]{\ensuremath{\mathsf{#1}}}
\renewcommand{\vec}[1]{\ensuremath{\bm{#1}}}

\newcommand{\norm}[1]{\ensuremath{\left\| #1\right\|}}
\newcommand{\abs}[1]{\ensuremath{\left| #1 \right|}}
\newcommand{\set}[1]{\ensuremath{\mathcal{#1}}}

\newcommand{\LGeoAttention}{\ensuremath{L^{\rm ga}}}
\newcommand{\LClassifAttention}{\ensuremath{L^{\rm ca}}}
\newcommand{\LGeoConf}{\ensuremath{L^{\rm gc}}}
\newcommand{\LClassifConf}{\ensuremath{L^{\rm cc}}}

\renewcommand{\th}{\ensuremath{\text{th}}}

\newcommand{\ra}[1]{\renewcommand{\arraystretch}{#1}}
\definecolor{LightGrey}{rgb}{.9,.9,.9}
\definecolor{White}{rgb}{1.,0.,1.}
\definecolor{first}{rgb}{.8,.0,.0}
\definecolor{second}{rgb}{.0,.6,.0}
\definecolor{third}{rgb}{.0,.0,.8}

\iccvfinalcopy 
\def\iccvPaperID{######} 

\ificcvfinal\fi

\begin{document}

\title{PCAM: Product of Cross-Attention Matrices\\for Rigid Registration of Point Clouds}

\author{Anh-Quan Cao\textsuperscript{2,}\thanks{Most of the work was done during an internship at valeo.ai in 2020.}
\and
Gilles Puy\textsuperscript{1}
\and 
Alexandre Boulch\textsuperscript{1}
\and
Renaud Marlet\textsuperscript{1, 3}
\and
\\
\textsuperscript{1}Valeo.ai, Paris, France \hspace{0.8cm}  
\textsuperscript{2}Inria, Paris, France\thanks{Inria, Mines ParisTech, PSL Research University.} \hspace{0.8cm} \\
\textsuperscript{3}LIGM, Ecole des Ponts, Univ Gustave Eiffel, CNRS, Marne-la-Vall\'ee, France
}

\maketitle

\ificcvfinal\thispagestyle{empty}\fi

\begin{abstract}
Rigid registration of point clouds with partial overlaps is a longstanding problem usually solved in two steps: (a) finding correspondences between the point clouds; (b) filtering these correspondences to keep only the most reliable ones to estimate the transformation. Recently, several deep nets have been proposed to solve these steps jointly. We built upon these works and propose PCAM: a neural network whose key element is a pointwise product of cross-attention matrices that permits to mix both low-level geometric and high-level contextual information to find point correspondences. These cross-attention matrices also permits the exchange of context information between the point clouds, at each layer, allowing the network construct better matching features within the overlapping regions. The experiments show that PCAM achieves state-of-the-art results among methods which, like us, solve steps (a) and (b) jointly via deepnets. Our code and trained models are available at \href{https://github.com/valeoai/PCAM}{https://github.com/valeoai/PCAM}.
\end{abstract}

%
\section{Introduction}

Point cloud registration is the problem of estimating the rigid transformation that aligns two point clouds. It has many applications in various domains such as autonomous driving, motion and pose estimation, 3D reconstruction, simultaneous localisation and mapping (SLAM), and augmented reality. The most famous method to solve this task is ICP \cite{besl92icp}, for which several improvements have been proposed, modifying the original optimisation process \cite{goicp,zhou16FGR} or using geometric feature descriptors \cite{SHOT} to match points.

Recently, end-to-end learning-based methods combining point feature extraction, point matching, and point-pairs filtering, have been developed to solve this task. Deep Closest Point (DCP) \cite{Wang2019DeepCP}, improved by PRNet \cite{Wang2019PRNetSL}, finds point correspondences via an attention matrix and estimate the transformation by solving a least-squares problem. Deep Global Registration (DGR) \cite{Choy2020DeepGR} tackles partial-to-partial point cloud registration by finding corresponding points via deep features, computing confidence scores for these pairs, and solving a weighted least-squares problem to estimate the transformation. IDAM \cite{IDAM} considers all possible pairs of points in a similarity matrix computed using deep or hand-crafted features, 
and proposes a learned two-step filter to select only relevant pairs for transformation estimation. Finally, \cite{3D-3D,StickyPillars,RPM-Net} use deep networks and the Sinkhorn algorithm to find and filter corresponding points.

Our observation is that, when matching points between point clouds, one would like to find correspondences using both local fine geometric information, to precisely select the best corresponding point, and high-level contextual information, to differentiate between points with similar local geometry but from different parts of the scene. The fine geometric information is naturally extracted at the first layers of a deep convnet, while the context information is found in the deepest layers. The existing deep registration methods estimate the correspondences between point clouds using features extracted at the deepest layers. Therefore, there is no explicit control on the amount of fine geometric and high-level context information encoded in these features.

Instead, we propose to compute point correspondences at every layer of our deep network via cross-attention matrices, and to combine these matrices via a point-wise multiplication. This simple yet very effective solution naturally ensures that both low-level geometric and high-level context information are exploited when matching points. It also permits to remove spurious matches found only at one scale. Furthermore, we also exploit these cross-attention matrices to exchange information between the point clouds at each layer, allowing the network to exploit context information from both point clouds to find the best matching point within the overlapping regions.

The design of our method is inspired by DGR \cite{Choy2020DeepGR}, DCP \cite{Wang2019DeepCP} and PRNet \cite{Wang2019PRNetSL}. We exploit cross-attention matrices to exchange context information between point clouds, like in \cite{Wang2019DeepCP,Wang2019PRNetSL,StickyPillars}, to find the best matches within overlapping region. Our main contribution is to propose to compute such cross-attention matrices at every network layer and to combine them to exploit both fine and high-level information when matching points. Our second contribution is to show that our method achieves state-of-the-art results on two real datasets (indoor, outdoor) and on a synthetic one.

%
\section{Related Work}

\textbf{Optimisation-based methods}. Iterative Closest Point (ICP) \cite{chen91icp,besl92icp,zhang94icp} is the most known algorithm for point cloud registration. It takes in two point clouds and alternate between point matching via nearest-neighbour search and transformation estimation by solving a least-squares problem. Several improvements of ICP's steps have been introduced to improve speed, solve the discretisation problem, become robust to outliers, or incorporate confidence scores on the correspondences \cite{agamennoni16PDA,bouaziz13sparseICP,pomerleau04review, rusinkiewicz01ICPvariants,segal09GenICP}. ICP often converges to a local minimum due to the non-convexity of the objective function. Therefore, some works propose solutions to widen the basin of attraction of the global optimum \cite{fitzgibbon02RobustReg,tsin2004}, use genetic algorithms \cite{silva05}, or find a good crude alignment to initialise ICP \cite{makadia06}. Another line of works concerns algorithms with theoretical convergence guarantees to a global optimum thanks to, \eg, the combination of ICP and a branch-and-bound technique \cite{goicp}, via convex relaxations \cite{maron10ConvexRelaxation,rosen20}, or thanks to mixed-integer programming \cite{Izatt17mixedInteger}. While theoretically appealing, these approaches usually suffers from a high computational complexity.

\textbf{Feature-based point matching}. Instead of relying solely on point coordinates to find correspondences between points, several approaches have proposed to extract point feature descriptors by analysing the local, and possibly global, 3D geometry. Classical methods use hand-crafted features such as \cite{spin, RANSAC, SHOT, shape_context} and, more recently, features extracted thanks to deep networks that take as inputs either hand-crafted local features \cite{Deng2018PPFFoldNet,Deng2018PPFNetGC,Gojcic2019ThePM, huang2020predator,Khoury2017LearningCG,zeng20163dmatch}, or directly point cloud coordinates \cite{Choy2019FullyCG,huang2020predator}. These methods use these point features to establish correspondences and rely on RANSAC to estimate the transformation, possibly using a pre-filtering of the points unlikely to provide good correspondences \cite{huang2020predator}.

\textbf{End-to-end learning.} Another type of approaches consists in training a network that will establish correspondences between the points of both point clouds, filter these correspondences to keep only the reliable pairs, and estimate the transformation based on the best pairs of points. Several methods fall in this category, such as \cite{Choy2020DeepGR, 3D-3D, StickyPillars, IDAM, Wang2019PRNetSL, Wang2019DeepCP, RPM-Net} which we described in the introduction. In addition, let us also mention PointNetLK \cite{PointNetLK} that extracts a global feature vector per point cloud and unroll the Lucas-Kanade algorithm \cite{LucasKanade} to find the best transformation aligning the point clouds in feature space, DeepGMR \cite{DeepGMR} that maps each point cloud to a parametric probability distribution via a neural network and estimates the transformation by minimising the KL-divergence between distributions. \cite{lawin2020registration} proposes a probabilistic registration method using deep features and learned attention weights. Finally, \cite{Gojcic_2020_CVPR} proposes a method for multiview registration by jointly training a pairwise registration module and a global refinement module.

%
\section{Network Architecture}

\begin{figure*}[t!]
\begin{center}
\includegraphics[width=\textwidth]{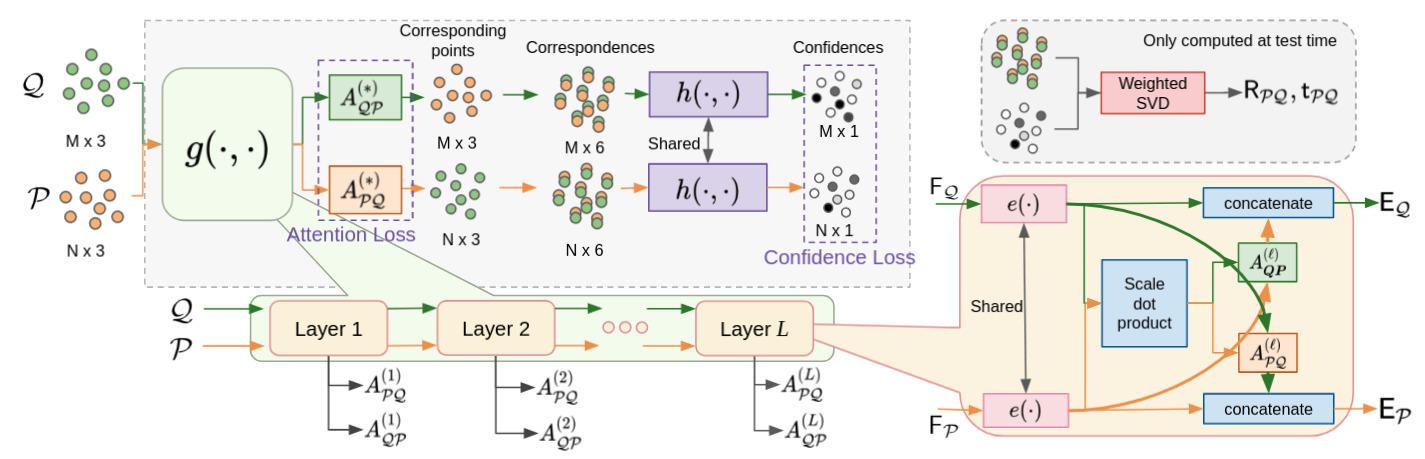}
\end{center}
\caption{\textbf{Overview of our architecture.} $\set{P}$ and $\set{Q}$ enter in the point matching module $g(\cdot)$ that permits the extraction of attention matrices used to compute one matching point for each input point. Each pair of matching points is then given a confidence score via $h(\cdot)$, which permits the estimation of the rigid transformation by solving a weighted least-squares problem via SVD.}
\label{fig:network-architecture}
\end{figure*}
%

\subsection{Problem Statement}

We consider the problem of partial-to-partial rigid registration between two point clouds $\set{P}=\{\vec{p}_1, \ldots, \vec{p}_{N}\}$ and $\set{Q} = \{\vec{q}_1, \ldots, \vec{q}_{M}\}$, with $\vec{p}_i, \vec{q}_j \in \mathbb{R}^3$. The point clouds $\set{P}$ and $\set{Q}$ represent two views of the same scene with partial overlap. Hence, only a subset of the points in $\set{P}$ can have matching points in $\set{Q}$, and vice versa from $\set{Q}$ to $\set{P}$. Let $\set{P}_\mathrm{v}$ (resp.\ $\set{Q}_\mathrm{v}$) be the subset of points in $\set{P}$ (resp.\ $\set{Q}$) visible in $\set{Q}$ (resp.\ $\set{P}$). Our goal is to estimate the rotation matrix $\ma{R}_{\rm gt} \in {\rm SO(3)}$ and translation vector $\vec{t}_{\rm gt} \in \Rbb^3$ that aligns $\set{P}_\mathrm{v}$ on $\set{Q}_\mathrm{v}$. This transformation can be estimated by solving
\begin{align}
\label{eq:icp_obj}
\min_{\ma{R}, \ma{t}} \sum_{\vec{p} \in \set{P}_\mathrm{v}}\norm{ \ma{R} \, \vec{p} + \vec{t} - m_{\set{Q}}(\vec{p})}_2^2,
\end{align}
where $m_{\set{Q}}: \set{P}_\mathrm{v} \rightarrow \set{Q}_\mathrm{v}$ maps any $\vec{p} \in \set{P}_\mathrm{v}$ to its best matching point $\vec{q} \in \set{Q}_\mathrm{v}$. Note that we can also estimate the inverse transformation from $\set{Q}$ to $\set{P}$ with the map $m_{\set{P}}\,{:}\, \set{Q}_\mathrm{v} \,{\rightarrow}\, \set{P}_\mathrm{v}$\rlap.

Problem \eqref{eq:icp_obj} can be solved via an SVD decomposition \cite{besl92icp,Proscutes}. The challenging tasks are the estimations of the subset of points $\set{P}_\mathrm{v}$ and $\set{Q}_\mathrm{v}$, and of the mappings $m_{\set{Q}}(\cdot)$ and $m_{\set{P}}(\cdot)$, given only $\set{P}$ and $\set{Q}$ as inputs.

\subsection{Method Overview}

Our method is composed of two classical modules: point matching and point-pair filtering (see Figure~\ref{fig:network-architecture}). The point-matching module, denoted by $g(\cdot, \cdot)$, permits us to estimate two maps $\tilde{m}_{\set{Q}}\,{:}\, \set{P} \,{\rightarrow}\, \set{Q}$ and $\tilde{m}_{\set{P}}\,{:}\, \set{Q} \,{\rightarrow}\, \set{P}$ that provide pairs of corresponding points between $\set{P}$ and $\set{Q}$ (even in non-overlapping regions). The point-pair filtering module, denoted by $h(\cdot, \cdot)$, provides a confidence score to all pairs $(\vec{p}_i, \tilde{m}_{\set{Q}}(\vec{p}_i))_{1\leq i\leq N}$, $(\vec{q}_j, \tilde{m}_{\set{P}}(\vec{q}_j))_{1\leq j\leq M}$, hence allowing detection of valid pairs in overlapping regions.

Our main contribution is in $g$, where we propose the construction of cross-attention matrices at each layer of $g$ and their combination to obtain point correspondences. These successive cross-attention matrices are obtained using point features with an increasing field of view, or scale, thus establishing point correspondences with increasing scene context. We obtain $\tilde{m}_{\set{Q}}$ and $\tilde{m}_{\set{P}}$ by combining all these attention matrices via pointwise matrix multiplications. Within overlapping regions, it permits us to improve the quality of the maps $\tilde{m}_{\set{Q}}$ and $\tilde{m}_{\set{P}}$ by filtering out the matches that are inconsistent across scales. We also use these cross-attention matrices to exchange information between $\set{P}$ and $\set{Q}$: each point in $\set{P}$ receives, in addition to its own feature, the feature of the best corresponding point in $\set{Q}$ (and vice-versa) before computing the point feature at the next layer. This process favours the discovery or rejection of correspondences at the next layer depending on the level of similarity between corresponding points at the current layer.

\subsection{First Module: Point Matching}
\label{sec:correspondences_estimation}

Our point matching module, denoted by $g\,{:}\,(\mathbb{R}^{N \times 3}, \, \mathbb{R}^{M \times 3}) \,{\xrightarrow[]{}}\, (\mathbb{R}^{N \times M}, \, \mathbb{R}^{N \times M})$, takes two unaligned point clouds $\set{P}$ and $\set{Q}$ as input, and produces two global attention matrices $\ma{A}^{(*)}_{\set{PQ}}$ and $\ma{A}^{(*)}_{\set{QP}}$ which inform us on the correspondences between point clouds. These global attention matrices are constructed (see Sec.~\ref{sec:global_attention}) from $L$ attention matrices, $\ma{A}^{(\ell)}_{\set{PQ}}$ and $\ma{A}^{(\ell)}_{\set{QP}}$, obtained at layers $\ell = 1, \ldots, L$ of our deep network (see Sec.~\ref{sec:layerwise_attention}).

\subsubsection{Layer-wise Cross-Attention Matrices $\ma{A}^{(\ell)}_{\set{PQ}}$, $\ma{A}^{(\ell)}_{\set{QP}}$}
\label{sec:layerwise_attention}

Let $\ma{F}_{\set{P}} \,{\in}\, \Rbb^{N \times c^{(\ell - 1)}}$ and $\ma{F}_{\set{Q}} \,{\in}\, \Rbb^{M \times c^{(\ell - 1)}}$ be the features on $\set{P}$ and $\set{Q}$, respectively, at the input of layer~$\ell$, where $c^{(\ell - 1)}$ is the size of the feature vectors at the output of layer $\ell\,{-}\, 1$.
Each point is thus described by a $c^{(\ell - 1)}$-dimensional feature vector. These features enter the $\ell^\th$ point convnet $e: \mathbb{R}^{\cdot \times c^{(\ell - 1)}} \xrightarrow[]{} \mathbb{R}^{\cdot \times c^{(\ell)}/2}$ to extract new features of size $c^{(\ell)}$/2 for each input point, to be doubled after concatenation of the best corresponding point feature in the other point cloud (see below). This encoder extracts local geometry information around each point in the point cloud thanks to three residual blocks, each containing two FKAConv \cite{boulch2020lightconvpoint} sub-layers. The supplementary material details $e(\cdot)$.

The new feature vectors $e(\ma{F}_{\set{P}})$ and $e(\ma{F}_{\set{Q}})$ are used to compute two attention matrices $\ma{A}^{(\ell)}_{\set{PQ}}, \ma{A}^{(\ell)}_{\set{QP}} \in \Rbb^{N \times M}$ with
\begin{align}
\label{eq:attn}
(\ma{A}^{(\ell)}_{\set{PQ}})_{ij} 
&= 
\frac
{\ee^{\, a_{ij} / s}}
{\sum_{k=1}^{M} \ee^{\, a_{ik} / s}}, \;
(\ma{A}^{(\ell)}_{\set{QP}})_{ij} 
= 
\frac
{\ee^{\, a_{ij} / s}}
{\sum_{k=1}^{N} \ee^{\, a_{kj} / s}},
\end{align}
where $s>0$ is the softmax temperature and 
\begin{align}
a_{ij} &= 
\frac
{e(\ma{F}_{\set{P}})_i \;
e(\ma{F}_{\set{Q}})_j^\adjoint
}
{
\|e(\ma{F}_{\set{P}})_i\|_2 \|e(\ma{F}_{\set{Q}})_j\|_2
}.
\end{align}
$\ma{A}_{\set{PQ}}, \ma{A}_{\set{QP}}$ differ in the normalisation dimension: 
$\ma{A}_{\set{PQ}}$ transfers information from $\set{Q}$ to $\set{P}$,
$\ma{A}_{\set{QP}}$ from $\set{P}$ to $\set{Q}$\rlap.

Finally, we transfer information between point clouds by computing
$
\ma{E}_{\set{P}} 
= 
[e(\ma{F}_{\set{P}}), \; \ma{A}_{\set{PQ}}^{(\ell)} \, e(\ma{F}_{\set{Q}})] \,{\in}\, \Rbb^{N \times c^{(\ell)}},
\ma{E}_{\set{Q}} 
= 
[e(\ma{F}_{\set{Q}}), \; {\ma{A}_{\set{QP}}^{(\ell)}}^\adjoint \, e(\ma{F}_{\set{P}})] \,{\in}\, \Rbb^{M \times c^{(\ell)}}.
$
$\ma{E}_{\set{P}}, \ma{E}_{\set{Q}}$ are the output features of layer $\ell$ and input features of layer $(\ell+1)$. At the end of the process, we have extracted two sets of $L$ attention matrices: $(\ma{A}^{(1)}_{\set{PQ}}, \ldots, \ma{A}^{(L)}_{\set{PQ}})$ and $(\ma{A}^{(1)}_{\set{QP}}, \ldots, \ma{A}^{(L)}_{\set{QP}})$.

\subsubsection{Global Attention Matrices $\ma{A}^{(*)}_{\set{PQ}}$, $\ma{A}^{(*)}_{\set{QP}}$}
\label{sec:global_attention}

We combine the attention matrices of each layer $\ell$ via a simple pointwise multiplication, denoted by $\odot$, to obtain
\begin{align}
\label{eq:product_attention}
\ma{A}^{(*)}_{\set{PQ}} = A^{(1)}_{\set{PQ}} \odot \ldots \odot A^{(L)}_{\set{PQ}}.
\end{align}
$\ma{A}^{(*)}_{\set{QP}}$ is defined similarly. The motivation for this strategy is that the successive attention matrices are constructed using features with an increasing field of view or scale, and we want to match points only if their features are similar at \emph{all} scales, hence the entry-wise multiplication of the attention coefficients. Another motivation is to permit to backpropagate gradients directly from the loss to each layer $\ell$.

\subsubsection{Soft and Sparse Maps}

There exists two standards ways to match points once the global attention matrices are available: soft and sparse mappings.
We explore the performance of both approaches in this work. The soft map $\tilde{m}_{\set{Q}}(\cdot)$ is defined as
\begin{align}
\label{eq:soft_assignement}
\tilde{m}_{\set{Q}}(\vec{p}_i) =  \frac{\sum_{j=1}^M (\ma{A}^{(*)}_{\set{PQ}})_{i,j} \; \vec{q}_j}
{\sum_{k=1}^M {(\ma{A}^{(*)}_{\set{PQ}})_{i,k}}}
\end{align}
for points $(\vec{p}_i)_{1\leq i \leq N}$. It maps $\set{P}$ to $\mathbb{R}^3$ rather than $\set{Q}$. The sparse map $\tilde{m}_{\set{Q}}(\cdot)$ is defined as
\begin{align}
\label{eq:sparse_assignement}
\tilde{m}_{\set{Q}}(\vec{p}_i) =  \vec{q}_{j^*},
\text{ where }
j^* = \argmax_j (\ma{A}^{(*)}_{\set{PQ}})_{ij}.
\end{align}
It maps $\set{P}$ to $\set{Q}$, but it is not differentiable and it prevents backpropagation from the confidence network to the point-matching network. Yet, the point-matching network will still be trained using a cross-entropy loss on the attention matrices (Sec.~\ref{sec:losses}).
The mapping $\tilde{m}_{\set{P}}$ from $\set{Q}$ to $\set{P}$ or $\mathbb{R}^3$ is constructed similarly using $\ma{A}^{(*)}_{\set{QP}}$.

\subsection{Second Module: Confidence Estimation}
\label{sec:confidence_estimation}

The module in Sec.\,\ref{sec:correspondences_estimation} produces pairs of matching points $(\vec{p}_i, \tilde{m}_{\set{Q}}(\vec{p}_i))$ and $(\vec{q}_j, \tilde{m}_{\set{P}}(\vec{q}_j))$ for all points in $\set{P}$ and $\set{Q}$. However, as we are tackling partial-to-partial registration, only subsets of $\set{P}$ and $\set{Q}$ match to one another. Hence, there are many incorrect matches which need to be filtered out. We detect these incorrect matches by concatenating each pair in a vector $[\vec{p}_i^\adjoint, \tilde{m}_{\set{Q}}(\vec{p}_i)^\adjoint]^\adjoint \in \mathbb{R}^6$, and then pass these $N$ pairs into the confidence estimation module $h: \mathbb{R}^{\cdot \times 6} \xrightarrow{} (0, 1)^{\cdot \times 1}$ which outputs a confidence score $w_{\vec{p}_i}$ for each input pair. The same module $h$ is used for pairs $(\vec{q}_j, \tilde{m}_{\set{P}}(\vec{q}_j))$ and yields $M$ corresponding scores $w_{\vec{q}_j}$. The confidence estimator is a point convnet with $9$ residual blocks, each containing $2$ FKAConv sub-layers. The confidence score are thus estimated using context information provided by the point convolution on $\mathcal{P}$ (or $\mathcal{Q}$). The detailed network architecture is in the supplementary material.

\subsection{Transformation Estimation}
\label{sec:weighted_svd}

We estimate the rotation and translation from $\set{P}$ to $\set{Q}$ by solving a weighted least-squares problem, using the weights $w_{\vec{p_i}}$ given by the confidence estimator:
\begin{align}
\label{eq:weighted_svd}
(\ma{R}_{\rm est}, \vec{t}_{\rm est}) = \argmin_{\ma{R}, \vec{t}} \sum_{i=1}^N \phi(w_{\vec{p_i}}) \norm{ \ma{R} \vec{p}_i + \vec{t} - \tilde{m}_{\set{Q}}(\vec{p}_i)}_2^2\rlap.
\end{align}
The function $\phi : \Rbb \xrightarrow{} \Rbb$ applies hard-thresholding on the confidence scores by setting to zero the less confident ones. The above problem can be solved with an SVD \cite{Proscutes,Wang2019DeepCP}.

\subsection{Losses}
\label{sec:losses}

The networks $g(\cdot, \cdot)$ and $h(\cdot, \cdot)$ and trained under supervision using a first loss, denoted by $\LClassifAttention$ that applies on the attention matrices $\ma{A}_{\set{PQ}}^{(*)}, \ma{A}_{\set{QP}}^{(*)}$ and the sum of other losses, $\LClassifConf$ and $\LGeoConf$, that apply on the confidence scores ${w}_{\vec{p}_i}, {w}_{\vec{q}_j}$. The complete training loss satisfies $\LClassifAttention + \LClassifConf + \LGeoConf$.

\subsubsection{Loss on the Attention Matrices}

The role of the attention matrices $\ma{A}_{\set{PQ}}^{(*)}, \ma{A}_{\set{QP}}^{(*)}$ is to identify mappings between $\set{P}$ and $\set{Q}$ for the final registration task. The pairs of points identified at this stage should include, as a subset, the ideal pairs encoded by $m_{\set{P}}$ and $m_{\set{Q}}$. Identifying this subset is the role of the confidence estimator $h(\cdot, \cdot)$.

The ideal map $m_{\set{P}}$ and $m_{\set{Q}}$ are estimated as follows. For each point $\vec{p}_i$, we search the closest point $\vec{q}_{j^*(i)}$ to $\ma{R}_{\rm gt}\vec{p}_i + \vec{t}_{\rm gt}$ in $\set{Q}$ and consider the pair $(\vec{p}_i, \vec{q}_{j^*(i)})$ as valid if $\ma{R}_{\rm gt}\vec{p}_i + \vec{t}_{\rm gt}$ is also the closest point to $\vec{q}_{j^*(i)}$ in $\{\ma{R}_{\rm gt}\vec{p}_u + \vec{t}_{\rm gt}\}_u$. More precisely, let us define
$
j^*(u) = \argmin_j \norm{\ma{R}_{\rm gt}\vec{p}_u + \vec{t}_{\rm gt} - \vec{q}_j}_2
$
and
$
i^*(v) = \argmin_v \norm{\ma{R}_{\rm gt}\vec{p}_i + \vec{t}_{\rm gt} - \vec{q}_v}_2,
$
The ideal maps satisfy
$
m_{\set{Q}}(\vec{p}_u) = \vec{q}_v,
\;
m_{\set{P}}(\vec{q}_v) = \vec{p}_u,
\;
\forall (u, v) \in \set{C},
$
where
$ 
\mathcal{C} = \{
(u, v) 
\; \vert \; 
j^*(u)=v, u=i^*(v) 
\}.
$
Only a subset of the points are in $\mathcal{C}$, even in overlapping regions. For convenience, we also define the set of points in $\set{P}$ for which a corresponding point exists in $\set{Q}$ by
$\mathcal{C}_{\set{P}} = \{u \; \vert \; \exists \, v \in \llbracket M \rrbracket \text{ s.t. } (u, v) \in \set{C} \}.$
The set $\mathcal{C}_{\set{Q}}$ is defined similarly for the inverse mapping.

We consider a contrastive classification loss, such as used in \cite{Choy2020DeepGR,IDAM,3D-3D}, denoted by $\LClassifAttention = \LClassifAttention_{\set{PQ}} + \LClassifAttention_{\set{QP}}$. $\LClassifAttention_{\set{PQ}}$ enforces a good mapping from $\set{P}$ to $\set{Q}$ and satisfies
\begin{align}
\LClassifAttention_{\set{PQ}}
&= - \inv{N} \sum_{(u, v) \in \set{C}} \log\left[(\ma{A}^{(*)}_{\set{PQ}})_{uv}\right]\nonumber\\
&
= - \inv{N} \sum_{\ell=1}^{L} \sum_{(u, v) \in \mathcal{C}}
\log\left[\frac{{\rm e}^{\, (a^{(\ell)}_{uv} / s)}}{\sum_{j=1}^{M} {\rm e}^{\, (a^{(\ell)}_{uj} / s)}} \right].
\end{align}
$\LClassifAttention_{\set{QP}}$ is defined likewise.
All points are constrained in (8) thanks to the softmax involved in the attention matrices.
In cases where $\mathcal{C}$ contains only few points, rather than augmenting $\mathcal{C}$ with pairs of worse quality, we leave $\mathcal{C}$ untouched to force the point matching network to identify the most reliable pairs (along with bad ones) and let the second module learn how to filter the bad pairs.

\subsubsection{Losses on the Confidence Scores}

To train the confidence estimator $h(\cdot, \cdot)$, we consider classification and geometric losses.

The classification losses are built by measuring the quality of the maps $\tilde{m}_{\set{Q}}, \tilde{m}_{\set{P}}$ at each input point. If a mapped point $\tilde{m}_{\set{Q}}(\vec{p}_i)$ is close to the ideal target point $\ma{R}_{\rm gt} \vec{p_i} + \vec{t}_{\rm gt}$, then the pair $(\vec{p}_i, \tilde{m}_{\set{Q}}(\vec{p}_i))$ is considered accurate and $w_{\vec{p}_i}$ should be close to $1$, and $0$ otherwise. The classification loss thus satisfies $\LClassifConf = \LClassifConf_{\set{PQ}} + \LClassifConf_{\set{QP}}$, where
\small
\begin{align*}
\LClassifConf_{\set{PQ}}
= 
-\bigg ( \inv{N} \sum_{i} y_{\vec{p}_i} \log(w_{\vec{p}_i}) 
+ (1 - y_{\vec{p}_i} ) \log (1 - w_{\vec{p}_i}) \bigg ),
\end{align*}
\normalsize
\begin{align}
\label{eq:threshold_classif_confidence}
y_{\vec{p}_i}
= 
\begin{cases}
1,  & \text{if } \|\ma{R}_{\rm gt} \vec{p_i} + \vec{t}_{\rm gt} - \tilde{m}_{\set{Q}}(\vec{p}_i)\|_2 \leq \kappa,\\
0,  & \text{otherwise},
\end{cases}
\end{align}
and $\kappa>0$ is a threshold to decide if a pair of points is accurate or not. The second loss $\LClassifConf_{\set{QP}}$ is defined similarly by considering the inverse mapping.

The threshold $\kappa>0$ to decide pair accuracy is somewhat arbitrary. To mitigate the effect of this choice, we also consider the geometric loss $\LGeoConf = \LGeoConf_{\set{PQ}} + \LGeoConf_{\set{QP}}$ where
\begin{align}
\label{eq:loss_geo_conf}
\LGeoConf_{\set{PQ}}
= \sum_i \frac{w_{\vec{p}_i}}{N} \; \|\ma{R}_{\rm gt} \vec{p_i} + \vec{t}_{\rm gt} - \tilde{m}_{\set{Q}}(\vec{p}_i)\|_2.
\end{align}
$\LGeoConf_{\set{QP}}$ is defined using $w_{\vec{q}_j}$ and the inverse transformation. Whenever the distance between a mapped point and its ideal location is large, then $w_{\vec{p}_i}$ should be small. On the contrary, when this distance is small, then $w_{\vec{p}_i}$ can be large. Note that this geometric losses can be used only in combination with the classification losses $\LClassifConf$ as otherwise $w_{\vec{p}_i}, w_{\vec{q}_j} = 0$ is a trivial useless solution.

%
\section{Experiments}

\subsection{Datasets and Training Parameters}

We evaluate our method on real (indoor, outdoor) and synthetic datasets. The indoor dataset is 3DMatch \cite{zeng20163dmatch}. We use the standard train/test splits and the procedure of \cite{Choy2020DeepGR,Deng2018PPFFoldNet, Deng2018PPFNetGC,Choy2019FullyCG} to generate pairs of scans with at least 30\% of overlap for training and testing. During training, as in \cite{Choy2020DeepGR}, we apply data augmentation using random rotations in $[0^{\circ}, 360^{\circ})$ around a random axis, and random scalings in $[0.8, 1.2]$. For the experiments on outdoor data, we use the KITTI odometry dataset \cite{KittiOdometry} and follow the same protocol as \cite{Choy2020DeepGR}: GPS-IMU is used to create pairs of scans that are at least 10m apart; the ground-truth transformation is computed using GPS followed by ICP. Unlike in \cite{Choy2020DeepGR}, we do not use data augmentation during training on KITTI. For synthetic data, we use ModelNet40 \cite{modelnet40} and follow the setup of \cite{Wang2019PRNetSL} to simulate partial registration problems.

All models are trained using AdamW \cite{adamw}, with a weight decay of $0.001$, a batch size of $1$, and a learning rate of $0.001$. On 3Dmatch and KITTI, the models are trained for $100$ epochs with a learning rate divided by $10$ after $60$ and $80$ epochs. On ModelNet40, it is sufficient to train the models for $10$ epochs (with a learning rate divided by $10$ after $6$ and $8$ epochs) to observe convergence. All results are reported using the models obtained at the last epoch. The temperature $s$ in \eqref{eq:attn} is set to $s=0.03$.

\subsection{Metrics}

The performance on 3DMatch and KITTI is measured using the metrics of \cite{Choy2020DeepGR}: the translation error (${\rm TE}$) and rotation error (${\rm RE}$) are defined as ${\rm TE}(\vec{t}_{\rm est}) = ||\vec{t}_{\rm est} - \vec{t}_{\rm gt}||_2$ and ${\rm RE}(\ma{R}_{\rm est}) = {\rm acos} \left[({\rm Tr}(\ma{R}^\adjoint_{\rm gt}   \ma{R}_{\rm est})-1)/2\right]$, respectively. 
We also computed the metric coined `recall' in \cite{Choy2020DeepGR}, which is the percentage of registrations whose rotation and translation errors are both smaller than predefined thresholds, i.e., the proportion of successful registrations.
We report the average rotation error $\rm RE_{all}$ and translation error $\rm TE_{all}$ on \emph{all} pairs of scans, as well as average rotation error ${\rm RE}$ and translation error ${\rm TE}$ on the subset of \emph{successful} registrations.
Performance on ModelNet40 is computed using the metrics of \cite{Wang2019PRNetSL}.

\begin{table}[t]
\begin{center}
\small
\ra{1.1}
\setlength{\tabcolsep}{2.5pt}
\begin{tabular}{@{}l| c c c | r r r r r@{}}
\toprule
& $\phi$
    & Opt.
    & Saf.
& $\rm TE_{all}$
    & $\rm RE_{all}$
    & Recall
    & $\rm TE$
    & $\rm RE$
\\
\midrule
FGR \cite{zhou16FGR}
&
    &
    &
& 
    & 
    & 42.7 
    & 0.11 
    & 4.08 
\\
RANSAC \cite{RANSAC} 
&
    &
    &
&
    &
    & 74.9
    & 0.09
    & 2.92 
\\
\midrule
\midrule
DCP \cite{Wang2019DeepCP} 
&
    &
    &
&
    &
    & 3.2
    & 0.21
    & 8.42 
\\
PointNetLK \cite{PointNetLK} 
&
    &
    &
&
    &
    & 1.6
    & 0.21  
    & 8.04
\\
\midrule
DGR \cite{Choy2020DeepGR}
& 
    & 
    &
& 0.47
    & 17.4
    & 73.9
    & 0.09
    & 2.97
\\
DGR \cite{Choy2020DeepGR}
& \checkmark
    & 
    &
& 0.38
    & 13.4
    & 81.5
    & 0.08
    & 2.56
\\
DGR \cite{Choy2020DeepGR}
& \checkmark
    & \checkmark
    &
& 0.33
    & 11.8
    & 86.6
    & \bf 0.07
    & 2.34
\\
DGR \cite{Choy2020DeepGR}
& \checkmark
    & \checkmark
    & \checkmark
& 0.25
    & 9.5
    & 91.2
    & \bf 0.07
    & 2.42
\\
\midrule
PCAM-Sparse
& 
    &
    &
& 0.44
    & 16.3
    & 74.9
    & 0.08
    & 2.98
\\
PCAM-Sparse
& \checkmark
    &
    &
& 0.35
    & 13.0
    & 83.8
    & \bf 0.07
    & 2.43
\\
PCAM-Sparse
& \checkmark
    & \checkmark
    & 
& 0.32
    & 11.8
    & 87.0
    & \bf 0.07
    & \bf 2.11
\\
PCAM-Sparse
& \checkmark
    & \checkmark
    & \checkmark
& \bf 0.23
    & \bf 8.9
    & \bf 92.4
    & \bf 0.07
    & 2.16
\\
\midrule
PCAM-Soft
& 
    &
    &
& 0.46
    & 17.5
    & 74.7
    & 0.09
    & 3.01
\\
PCAM-Soft
& \checkmark
    &
    &
& 0.37
    & 13.1
    & 81.4
    & 0.08
    & 2.50
\\
PCAM-Soft
& \checkmark
    & \checkmark
    & 
& 0.33
    & 11.9
    & 85.6
    & \bf 0.07
    & 2.12
\\
PCAM-Soft
& \checkmark
    & \checkmark
    & \checkmark
& 0.24
    & 9.8
    & 91.3
    & \bf 0.07
    & 2.25
\\
\bottomrule
\end{tabular}
\end{center}
\caption{\textbf{Results on 3DMatch test split}. The recall is computed using $0.3$ m and $15^\circ$ as thresholds.  The scores of all variants of DGR are obtained using the official implementation of DGR \cite{Choy2020DeepGR}, and those of the other concurrent methods are reported from \cite{Choy2020DeepGR}. The columns `$\phi$', `Opt.', and `Saf.' indicate respectively whether hard-thresholding on the confidence scores, the pose refinement of \cite{Choy2020DeepGR}, and the safeguard registration of \cite{Choy2020DeepGR}, are used or not.}
\label{tab:3dmatch}
\end{table}
%

\subsection{Comparison with Existing Methods}

We report in this section the performance of PCAM on 3DMatch and KITTI. The results obtained on ModelNet40 are available in the supplementary material and show that PCAM outperforms the concurrent methods on this dataset. In what follows, PCAM-Soft and PCAM-Sparse refer to our method with soft \eqref{eq:soft_assignement} and sparse maps \eqref{eq:sparse_assignement}, respectively. 

\subsubsection{Indoor Dataset: 3DMatch}
\label{sec:3dmatch}

We train PCAM on 3DMatch using $L=6$ layers, with point clouds obtained by sampling $N=M=4096$ points at random from the voxelised point clouds (voxel size of 5 cm). The parameter $\kappa$ in \eqref{eq:threshold_classif_confidence} is set at $12$ cm.

For a thorough comparison with DGR \cite{Choy2020DeepGR}, we study the performance of PCAM after applying each DGR's post-processing steps: filtering the confidence weights with $\phi$, refining the pose estimation using the optimisation proposed in \cite{Choy2020DeepGR}, and using the same safeguard as in \cite{Choy2020DeepGR}. The threshold applied in $\phi$ is tuned on 3DMatch's validation set. The safeguard is activated when the $\ell_1$-norm of the confidence weights is below a threshold, in which case the transformation is estimated using RANSAC (see details in \cite{Choy2020DeepGR}). For a fair comparison with DGR, we compute this second threshold such that DGR and PCAM use the safeguard on the same proportion of scans.

We report the performance of PCAM and concurrent methods in Table~\ref{tab:3dmatch}. PCAM achieves the best results and this is confirmed in the curves of Fig.~\ref{fig:registrations_result}. We provide illustrations of the quality of matched points and registrations in the supplementary material. We did not compare PCAM with RLL, which is not performing as well as DGR on 3DMatch~\cite{lawin2020registration}.

\subsubsection{Outdoor Dataset: KITTI}

\begin{table}[t]
\begin{center}
\small
\ra{1.1}
\setlength{\tabcolsep}{2.5pt}
\begin{tabular}{@{}l| c c c | r r r r r@{}}
\toprule
& $\phi$
    & Opt.
    & ICP
& $\rm TE_{all}$
    & $\rm RE_{all}$
    & Recall
    & $\rm TE$
    & $\rm RE$
\\
\midrule
FGR \cite{zhou16FGR}
&
    &
    &
& 
    & 
    & 0.2
    & 0.41 
    & 1.02 
\\
RANSAC \cite{RANSAC} 
&
    &
    &
&
    &
    & 34.2
    & 0.26
    & 1.39
\\
FCGF \cite{Choy2019FullyCG} 
&
    &
    &
&
    &
    & \bf 98.2
    & 0.1
    & 0.33
\\
\midrule
\midrule
DGR \cite{Choy2020DeepGR}
& 
    &
 
    &
& 0.77
    & 2.32
    & 63.1
    & 0.28
    & 0.56
\\
DGR \cite{Choy2020DeepGR}
& \checkmark
    & 
    &
& 0.76
    & 2.32
    & 63.4
    & 0.28
    & 0.56
\\
DGR \cite{Choy2020DeepGR}
& \checkmark
    & \checkmark
    & 
& 0.34
    & 1.62
    & 96.6
    & 0.21
    & 0.33
\\
DGR \cite{Choy2020DeepGR}
& \checkmark
    & \checkmark
    & \checkmark
& 0.16
    & 1.43
    & \bf 98.2
    & \bf 0.03
    & \bf 0.14
\\
\midrule
PCAM - Soft
& 
    &
    & 
& 0.18
    & 1.00
    & 97.2
    & 0.08
    & 0.33
\\
PCAM - Sparse
& 
    &
    &
& 0.22
    & 1.17
    & 96.5
    & 0.08
    & 0.31
\\
PCAM - Soft
& 
    &
    & \checkmark
& \bf 0.12
    & \bf 0.79
    & 98.0
    & \bf 0.03
    & \bf 0.14
\\
PCAM - Sparse 
& 
    &
    & \checkmark
& 0.17
    & 1.04
    & 97.4
    & \bf 0.03
    & \bf 0.14
\\
\bottomrule
\end{tabular}
\end{center}
\caption{\textbf{Results on KITTI test split}. The recall is computed using $0.6$ m and $5^\circ$ as thresholds. The scores of all variants of DGR are obtained using the official implementation of DGR \cite{Choy2020DeepGR}, and those of the other concurrent methods are reported from \cite{Choy2020DeepGR}. The columns `$\phi$', `Opt.', and `ICP' indicate respectively whether hard-thresholding on the confidence scores, the pose refinement of \cite{Choy2020DeepGR}, and post-processing by ICP, are used or not.
}
\label{tab:kitti}
\vspace*{-5mm}
\end{table}

We train PCAM on 3DMatch using $L=6$ layers with point clouds of (at most) $N=M=2048$ points drawn at random from the voxelised point clouds (voxel size of 30 cm). The parameter $\kappa$ in \eqref{eq:threshold_classif_confidence} is set at $60$ cm.

\begin{figure*}[t]
\centering
\includegraphics[width=\linewidth,trim=40 5 40 16,clip]{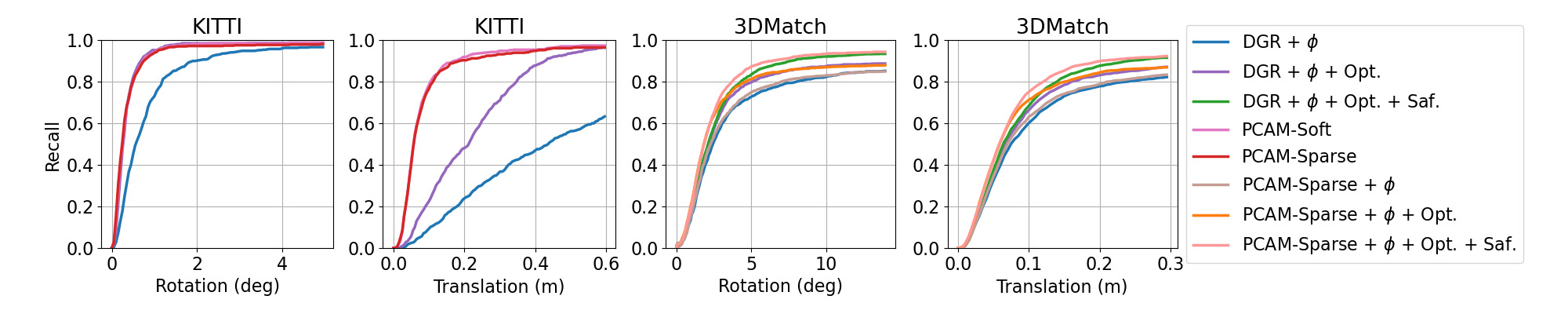}
\caption{Registration recall vs rotation and translation error thresholds for KITTI (left) and 3DMatch (right).}
\label{fig:registrations_result}
\end{figure*}

The performance of our method is compared to others in Table~\ref{tab:kitti}. As scores after refinement of DGR registration by ICP are reported on this dataset in \cite{Choy2020DeepGR}, we also include the scores of PCAM after the same refinement in this table. PCAM outperforms DGR on all metrics with both sparse \eqref{eq:sparse_assignement} and soft \eqref{eq:soft_assignement} maps, except on the recall when using sparse maps, where it is just $0.1$ point behind DGR. When combined with ICP, PCAM also achieves better results than DGR except on the recall but where it is just 0.2 point behind with soft maps and 0.8 with sparse maps. Finally, we notice in Fig.~\ref{fig:registrations_result} a clear advantage of PCAM over DGR for the estimation of the translation. Note that PCAM achieves better performance than DGR without the need to use $\phi$, nor used the pose refinement of \cite{Choy2020DeepGR}; we did not notice any significant improvement when using them on KITTI. We compare PCAM to RLL on the version of KITTI used in \cite{lawin2020registration}: PCAM outperforms RLL with a recall of $84.7\%$ for sparse maps and $86.5\%$ for soft maps, vs $76.9\%$ for RLL.

\begin{table*}[t]
\begin{center}
\small
\ra{1.1}
\setlength{\tabcolsep}{3pt}
\begin{tabular}{@{} c c c c | c c c c | c c c | c c c | c c @{}}
\toprule
    & \multicolumn{3}{c}{}
    & \multicolumn{4}{c|}{}
    & \multicolumn{3}{c|}{3DMatch}
    & \multicolumn{3}{c|}{KITTI}
    & \multicolumn{2}{c}{ModelNet40}
\\
\midrule
    & $\LClassifAttention$ 
    & $\LGeoAttention$
    & $\LGeoConf$
& $\odot_\ell \; \ma{A}^{(\ell)}$
    & $\ma{A}^{(L)}$
    & $\ma{A}$
    & L
& Rec. (\%)
    & $\rm RE_{all}$
    & $\rm TE_{all}$
& Rec. (\%)
    & $\rm RE_{all}$
    & $\rm TE_{all}$
& \parbox{15mm}{RMSE (R)\\($\times 10^{-3}$)}
    & \parbox{15mm}{RMSE (t)\\($\times 10^{-3}$)}
\\
\midrule
\parbox[t]{1mm}{\multirow{7}{*}{\rotatebox[origin=c]{90}
{
Soft map 
}}}
& \checkmark
    & \checkmark
    & \checkmark
& \checkmark
    & 
    &
    & 2
& 53.7 \scriptsize (1.8)
    & 36.4 \scriptsize (1.9)
    & 0.49 \scriptsize (0.03)
& 93.4 \scriptsize (0.6)
    & 3.0 \scriptsize (0.2)
    & 0.45 \scriptsize (0.02)
& 18 \scriptsize (1.2)
    & 0.13	\scriptsize (0.002)
\\
& \checkmark
    &
    & \checkmark
& \checkmark
    & 
    &
    & 2
& 52.3 \scriptsize (0.6)
    & 36.3 \scriptsize (1.0)
    & 0.49 \scriptsize (0.01)
& 94.3 \scriptsize (0.5)
    & 2.5 \scriptsize (0.1)
    & 0.42 \scriptsize (0.02)
& 15 \scriptsize (2.7)
    & 0.09	\scriptsize (0.014)
\\
& \checkmark
    & 
    & 
& \checkmark
    & 
    &
    & 2
& 45.9 \scriptsize (1.5)
    & 41.9 \scriptsize (2.5)
    & 0.55 \scriptsize (0.03)
& 93.7 \scriptsize (0.3)
    & 3.0 \scriptsize (0.1)
    & 0.44 \scriptsize (0.01)
& 18 \scriptsize (1.1)
    & 0.12	\scriptsize (0.015)
\\
&
    & \checkmark
    & \checkmark
& \checkmark
    & 
    &
    & 2
& 19.4 \scriptsize (2.7)
    & 67.4 \scriptsize (1.5)
    & 0.85 \scriptsize (0.02)
& 90.9 \scriptsize	(1.1)
    & 3.1 \scriptsize (0.1)
    & 0.52 \scriptsize (0.04)
& \llap494 \scriptsize (21\rlap{8)}\hphantom)
    & 1.78	\scriptsize (0.879)
\\
\cmidrule(lr){2-16}
& \checkmark
    & 
    & \checkmark
& 
    & \checkmark
    &
    & 2
& 37.1 \scriptsize (2.9) 
    & 47.9 \scriptsize (2.2)
    & 0.65 \scriptsize (0.02)
& 90.4 \scriptsize (0.9)
    & 3.5 \scriptsize (0.1)
    & 0.61 \scriptsize (0.08)
& 38 \scriptsize (3.9)
    & 0.28 \scriptsize (0.007)
\\
& \checkmark
    & 
    & \checkmark
& 
    & 
    & \checkmark
    & 2
& 37.3 \scriptsize (1.4) 
    & 47.2 \scriptsize (0.6)
    & 0.65 \scriptsize (0.01)
& 92.7 \scriptsize (0.4)
    & 2.8 \scriptsize (0.2)
    & 0.53 \scriptsize (0.01)
& 66 \scriptsize (10.\rlap{5)}\hphantom)
    & 0.39 \scriptsize (0.010)
\\
\cmidrule(lr){2-16}
& \checkmark
    & 
    & \checkmark
& \checkmark
    & 
    &
    & 6
& 82.4 \scriptsize (0.3)
    & 13.5 \scriptsize (0.3)
    & 0.21	\scriptsize (0.01)
& 95.4 \scriptsize (0.3)
    & 2.2 \scriptsize (0.1)
    & 0.33 \scriptsize (0.01)
& 16 \scriptsize \hphantom{(0.0)}
    & 0.11 \scriptsize \hphantom{(0.000)}
\\
& \checkmark
    & 
    & \checkmark
& 
    & \checkmark
    &
    & 6
& 76.6 \scriptsize (1.9)
    & 18.2 \scriptsize (0.8)
    & 0.27 \scriptsize (0.01)
& 93.5 \scriptsize (0.8)
    & 2.9 \scriptsize (0.03)
    & 0.46 \scriptsize (0.03)
& 22 \scriptsize \hphantom{(0.0)}
    & 0.16 \scriptsize \hphantom{(0.000)}
\\
& \checkmark
    & 
    & \checkmark
& 
    & 
    & \checkmark
    & 6
& 64.4 \scriptsize (1.1)
    & 27.8 \scriptsize (0.6)
    & 0.38 \scriptsize (0.01)
& 93.9	\scriptsize (0.0)
    & 2.0 \scriptsize (0.1)
    & 0.33 \scriptsize (0.01)
& 28 \scriptsize \hphantom{(0.0)}
    & 0.16 \scriptsize \hphantom{(0.000)}
\\
\midrule
\midrule
\parbox[t]{1mm}{\multirow{7}{*}
{\rotatebox[origin=c]{90} 
{
Sparse map 
}
}
}
& \checkmark
    & \scriptsize N/A
    & \checkmark
& \checkmark
    & 
    &
    & 2
& 52.4 \scriptsize (2.1)
    & 38.6 \scriptsize (2.0)
    & 0.51 \scriptsize (0.02)
& 94.6 \scriptsize (0.5)
    & 2.8 \scriptsize (0.5) 
    & 0.39 \scriptsize (0.02)
& 19 \scriptsize (1.9)
    & 0.10 \scriptsize (0.016)
\\
& \checkmark
    & \scriptsize N/A
    & 
& \checkmark
    & 
    &
    & 2
& 48.5 \scriptsize (1.0)
    & 39.4 \scriptsize (1.5)
    & 0.53 \scriptsize (0.02)
& 94.4 \scriptsize (0.4)
    & 2.6 \scriptsize (0.4)
    & 0.36 \scriptsize (0.03)
& 24 \scriptsize (6.6)
    & 0.14 \scriptsize (0.031)
\\
\cmidrule(lr){2-16}
& \checkmark
    & \scriptsize N/A
    & \checkmark
& 
    & \checkmark
    &
    & 2
& 54.1 \scriptsize (3.8)
    & 36.0 \scriptsize (2.7)
    & 0.49 \scriptsize (0.03)
& 94.2 \scriptsize (0.5)
    & 2.6 \scriptsize (0.2)
    & 0.43 \scriptsize (0.05)
& 15 \scriptsize (0.4)
    & 0.09 \scriptsize (0.002)
\\
& \checkmark
    & \scriptsize N/A
    & \checkmark
& 
    & 
    & \checkmark
    & 2
& 50.9 \scriptsize (1.3)
    & 38.9 \scriptsize (1.5)
    & 0.52 \scriptsize (0.02)
& 94.6 \scriptsize (0.8)
    & 2.7 \scriptsize (0.3)
    & 0.43 \scriptsize (0.03)
& 17 \scriptsize (1.3)
    & 0.09 \scriptsize (0.004)
\\
\cmidrule(lr){2-16}
& \checkmark
    & \scriptsize N/A
    & \checkmark
& \checkmark
    & 
    &
    & 6
& 81.6 \scriptsize (1.2)
    & 16.3 \scriptsize (1.2)
    & 0.24 \scriptsize (0.01)
& 95.9 \scriptsize (0.3)
    & 2.3 \scriptsize (0.05)
    & 0.33 \scriptsize (0.01)
& 36 \scriptsize\hphantom{(0.0)}
    & 0.20 \scriptsize\hphantom{ (0.000)}
\\
& \checkmark
    & \scriptsize N/A
    & \checkmark
& 
    & \checkmark
    &
    & 6
& 78.4 \scriptsize (1.1)
    & 17.5 \scriptsize (1.3)
    & 0.24 \scriptsize (0.01)
& 96.5 \scriptsize (0.3)
    & 2.4 \scriptsize (0.1)
    & 0.46 \scriptsize (0.003)
& 27 \scriptsize\hphantom{(0.0)}
    & 0.15 \scriptsize\hphantom{(0.000)}
\\
& \checkmark
    & \scriptsize N/A
    & \checkmark
& 
    & 
    & \checkmark
    & 6
& 68.0 \scriptsize (1.1)
    & 25.9 \scriptsize (0.5)
    & 0.34 \scriptsize (0.01)
& 95.7 \scriptsize (0.3)
    & 1.9 \scriptsize (0.1)
    & 0.33 \scriptsize (0.01)
& 18 \scriptsize\hphantom{(0.0)}
    & 0.10 \scriptsize\hphantom{(0.000)}
\\
\bottomrule
\end{tabular}
\end{center}
\caption{\textbf{Ablation study on the validation set of 3DMatch, KITTI, and ModelNet40}. The losses $\LClassifAttention$, $\LGeoConf$ $\LGeoAttention$ are defined in the main paper. The global attention matrix $\ma{A}^{(*)}$ is either equal to the pointwise multiplication of all the attention matrices ($\odot_\ell \ma{A}^{(\ell)}$), or to the last attention matrix ($\ma{A}^{(L)}$), or to the attention matrix ($\ma{A}$) computed at the last layer of our network in a version where we have removed all the intermediate attention matrices $\ell=1, \ldots, L$ (see main paper for details).}
\label{tab:ablation}
\end{table*}
%

\subsection{Ablation Study}

In this section, we conduct two ablation studies: the first justifies our choice of the training loss; the second demonstrates the benefit of the product of attention matrices and of the exchange of contextual information between point clouds. The performance of the trained models are evaluated on the validation set of the considered datasets. On 3DMatch and KITTI, the input point clouds are obtained by drawing at random (at most) $N=M=2048$ points from the voxelised point clouds. Because of these random draws, different evaluations lead to different scores. To take into account these variations, we report the average scores and the standard deviations obtained using three evaluations on the validation set for each trained model. On ModelNet40, we do not subsample the point clouds. The rotation $\ma{R}_{\rm est}$ and translation $\vec{t}_{\rm est}$ are computed using the method described in Sec.~\ref{sec:weighted_svd} \emph{without} using any hard-thresholding on the confidence scores. We conduct experiments with $L\,{=}\,2$ or $L\,{=}\,6$ layers. For the experiments at $L\,{=}\,2$, two models are trained from different initialisation and the reported average scores and the standard deviations are computed using both models. One model is trained for the experiments at $L\,{=}\,6$.

\subsubsection{Training Loss}
\label{sec:exp_training_losses}

The experiments in this section are conducted with $L\,{=}\,2$ layers and using the product of attention matrices \eqref{eq:product_attention}.

We recall that the loss $\LClassifConf$ cannot be removed as there exists a trivial solution where all the confidence scores are equal to zero in absence of this loss. Hence, the only loss that can be removed is $\LGeoConf$. The results in Table~\ref{tab:ablation} show that removing $\LGeoConf$ yields worse performance both with soft and sparse mappings. We explain this result by the fact that $\ma{R}_{\rm est}$ and $\vec{t}_{\rm est}$ are obtained using a slightly modified version of $\LGeoConf$ at test time (see Sec.~\ref{sec:weighted_svd}), hence the presence of $\LGeoConf$ at training time probably improves the quality of confidence scores for the estimation of the transformation.

With soft maps \eqref{eq:soft_assignement}, we can use another loss to train $g$ in replacement or in complement to $\LClassifAttention$. The motivation for this loss is that $\LClassifAttention$ penalises all points that are not mapped correctly to the ideal target point in the same way, whether the mapped point is close to the ideal location or far away. Instead, we can consider the possibility to use a geometric loss, $\LGeoAttention = \LGeoAttention_{\set{PQ}} + \LGeoAttention_{\set{QP}}$, that takes into account the distance between the estimated and ideal corresponding points:
\begin{align}
\LGeoAttention_{\set{PQ}}
& = 
\inv{\abs{\set{C}_{\set{P}}}} \; \sum_{u \in \set{C}_{\set{P}}} \norm{\tilde{m}_{\set{Q}}(\vec{p}_u) - m_{\set{Q}}(\vec{p}_u)}_2.
\end{align}
$\LGeoAttention_{\set{QP}}$ is defined likewise using $\set{C}_{\set{Q}}$. Note that it can be used to train $g$ only in the case of soft maps \eqref{eq:soft_assignement} as $\tilde{m}_{\set{Q}}$ is not differentiable when using sparse maps \eqref{eq:sparse_assignement}. The results in Table~\ref{tab:ablation} show that the performance drops significantly when using $\LGeoAttention$ alone. Note that a similar observation was made in \cite{zodage20}. Finally, using $\LGeoAttention + \LClassifAttention$ yields mildly better performance only on 3DMatch compared to using $\LClassifAttention$ alone.

In view of these results, we chose to train PCAM using $\LClassifAttention + \LClassifConf + \LGeoConf$, both for soft and sparse maps.

%

\subsubsection{Role of the Attention Matrices} 
\label{sec:ablation_product}

\textbf{Product of attention matrices: $\odot_\ell \ma{A}^{(\ell)}$ vs $\ma{A}^{(L)}$}. To show the benefit of mixing low-level geometric information and high-level context information via the proposed product of attention matrices \eqref{eq:product_attention}, we compare the performance of PCAM when replacing \eqref{eq:product_attention} by $\ma{A}^{(*)} = \ma{A}^{(L)}$, \ie, using only the cross-attention matrix at the deepest layer. Note that the matrices $\ma{A}^{(1)}, \ldots, \ma{A}^{(L-1)}$ are still present in $g$.

We expect the interest of the mix of information \eqref{eq:product_attention} to appear at large $L$ (so that there is enough information to mix) and on challenging datasets where small overlaps and large transformations yield many ambiguities to be resolved for accurate point matching. In that respect, the most challenging dataset used in this work is 3DMatch. One can indeed verify in Table~\ref{tab:ablation} that all models yields quite similar and very good results on KITTI and ModelNet40, leaving little room to clearly observe the impact of different network architectures. On ModelNet40, we highlight that all scores are close to zero (the scores are multiplied by $10^{3}$ in Table~\ref{tab:ablation}) which makes it difficult to rank the different strategies as the difference between them could be due to training noise. In contrast, the impact of different choices of architecture is much more visible on 3DMatch. 

We clearly observe the benefit of increasing $L$ from $2$ to~$6$, and of mixing low-level geometric information and high-level context via the product of cross-attention matrices at $L=6$ on 3DMatch. At $L\,{=}\,2$, we observe a benefit of the product of attention matrices when using soft maps on 3DMatch, while this advantage is not visible when using sparse maps. We explain this fact because the product can make the soft maps more tightly concentrated around the found corresponding points, yielding better quality maps. This effect is invisible with sparse maps at $L\,{=}\,2$ because, by construction, these maps are tightly concentrated around the corresponding points that are found.

\textbf{Sharing contextual information: $\ma{A}^{(L)}$ vs $\ma{A}$}. To show the benefit of exchanging information between point clouds via the intermediate cross-attention matrices $\ma{A}^{(1)}, \ldots, \ma{A}^{(L-1)}$, we conduct experiments where we remove all intermediate matrices but the last one, which we denote by $\ma{A}$. As before, we expect the benefit of this strategy to be more visible at large $L$ and on challenging datasets such as 3DMatch. The results in Table~\ref{tab:ablation} confirm this result where the benefit of exchanging contextual information is particularly noticeable at $L=6$ on 3DMatch.

%
\section{Discussion}

In this section we differentiate our method from the most closely related works on three main aspects: exchange of contextual information between point clouds, computation of the pairs of corresponding points, and point-pair filtering.

\textbf{Exchange of contextual information.} Multiple cross-attention layers in PCAM allows the network to  progressively exchange information between point clouds  at \emph{every} layer to find the best matches in overlapping regions. DCP \cite{Wang2019DeepCP}, PRNet \cite{Wang2019PRNetSL}, and OPRNet \cite{3D-3D} use transformer layers that also exchange information between point clouds, but only in the deepest layers of their convnet. PREDATOR \cite{huang2020predator} uses one cross-attention layer in the middle of their U-Net, hence only exchanging high-level contextual information. The newly published method \cite{StickyPillars} uses several cross-attention layers but, unlike us, does not merge them for point matching. There is no exchange of information between point clouds in \cite{Choy2020DeepGR, IDAM, RPM-Net}.

\textbf{Point matching.} PCAM uses \emph{multiple} cross-attention matrices that allows the network to exploit both fine-grained geometric information and high-level context to find pairs of corresponding points. In contrast, DCP  \cite{Wang2019DeepCP} and PRNet \cite{Wang2019PRNetSL} use only one attention layer at the deepest layer of their network, which might limit their performance due to the lack of this fine-grained geometric information. RPM-Net\cite{RPM-Net}, OPRNet \cite{3D-3D} use a single Sinkhorn layer to establish the correspondences. IDAM \cite{IDAM} replaces the dot product operator by a learned convolution module that computes the similarity score between point features. The point-matching module in \cite{Gojcic_2020_CVPR} is equivalent to ours in the non-optimal setup of one attention $\ma{A}$.

\textbf{Point filtering.} PCAM first constructs pairs of corresponding points and assign to each of them a confidence score using a point convolution network. A similar strategy is used in \cite{Gojcic_2020_CVPR}. DCP \cite{Wang2019DeepCP} does not have any filtering step, hence cannot handle partial-to-partial registration. Before starting to match points, PRNet \cite{Wang2019PRNetSL} first selects a subset of points to be matched in each point cloud using the $\ell_2$-norm of point features. Similarly to PRNet \cite{Wang2019PRNetSL}, PREDATOR \cite{huang2020predator} also start with a selection of points to be matched in each point cloud, using the learned overlap probability and matchability scores. DGR \cite{Choy2020DeepGR} uses a similar process to PCAM but uses 6D convolutions to predict the confidence score, whereas we use 3D convolutions with 6D point-features (the concatenation of the coordinates of the point itself and its match point). IDAM \cite{IDAM} proposes a two-stage point elimination technique where points are individually filtered first; pairs of points are then discarded. Finally, \cite{3D-3D, StickyPillars, RPM-Net} use their Sinkhorn layer to filter outliers.

%
\section{Conclusion}

In this work, we have proposed a novel network architecture where we transfer information between point clouds to find the best match within the overlapping regions, and mix low-level geometric information and high-level contextual knowledge to find correspondences between point clouds. Our experiments show that this architecture achieves state-of-the-art results on several datasets.

\vspace{1mm}
\noindent\textbf{Acknowledgements.} This work was partly performed using HPC resources from GENCI–IDRIS (Grant 2020-AD011012040). We thank Raoul de Charette for his constructive feedbacks on an earlier version of this paper.

%
{\small
\bibliographystyle{ieee_fullname}
\bibliography{main}

\begin{thebibliography}{10}\itemsep=-1pt

\bibitem{agamennoni16PDA}
G. {Agamennoni}, S. {Fontana}, R.~Y. {Siegwart}, and D.~G. {Sorrenti}.
\newblock {Point Clouds Registration with Probabilistic Data Association}.
\newblock In {\em International Conference on Intelligent Robots and Systems
  (IROS)}, pages 4092--4098, 2016.

\bibitem{PointNetLK}
Y. Aoki, H. Goforth, R.~A. Srivatsan, and S. Lucey.
\newblock {PointNetLK: Robust \& Efficient Point Cloud Registration Using
  PointNet}.
\newblock In {\em Conference on Computer Vision and Pattern Recognition
  (CVPR)}, pages 7156--7165, 2019.

\bibitem{besl92icp}
P.~J. {Besl} and N.~D. {McKay}.
\newblock {A method for registration of 3-D shapes}.
\newblock {\em IEEE Transactions on Pattern Analysis and Machine Intelligence},
  14(2):239--256, 1992.

\bibitem{bouaziz13sparseICP}
S. Bouaziz, A. Tagliasacchi, and M. Pauly.
\newblock {Sparse Iterative Closest Point}.
\newblock {\em Computer Graphics Forum}, 32(5):113--123, 2013.

\bibitem{boulch2020lightconvpoint}
A. Boulch, G. Puy, and R. Marlet.
\newblock {FKAConv: Feature-Kernel Alignment for Point Cloud Convolution}.
\newblock In {\em Asian Conference on Computer Vision (ACCV)}, 2020.

\bibitem{chen91icp}
Y. Chen and G. Medioni.
\newblock {Object modeling by registration of multiple range images}.
\newblock In {\em International Conference on Robotics and Automation (ICRA)},
  volume~3, pages 2724–--2729, 1991.

\bibitem{Choy2020DeepGR}
C. Choy, W. Dong, and V. Koltun.
\newblock {Deep Global Registration}.
\newblock In {\em Conference on Computer Vision and Pattern Recognition
  (CVPR)}, pages 2511--2520, 2020.

\bibitem{Choy2019FullyCG}
C. Choy, J. Park, and V. Koltun.
\newblock Fully convolutional geometric features.
\newblock In {\em International Conference on Computer Vision (ICCV)}, pages
  8957--8965, 2019.

\bibitem{3D-3D}
Z. Dang, F. Wang, and M. Salzmann.
\newblock {Learning 3D-3D Correspondences for One-Shot Partial-to-partial
  Registration}.
\newblock {\em arXiv:2006.04523}, 2020.

\bibitem{Deng2018PPFFoldNet}
H. Deng, T. Birdal, and S. Ilic.
\newblock {PPF-FoldNet: Unsupervised Learning of Rotation Invariant 3D Local
  Descriptors}.
\newblock In {\em European Conference on Computer Vision (ECCV)}, 2018.

\bibitem{Deng2018PPFNetGC}
H. Deng, T. Birdal, and S. Ilic.
\newblock {PPFNet: Global Context Aware Local Features for Robust 3D Point
  Matching}.
\newblock In {\em Conference on Computer Vision and Pattern Recognition
  (CVPR)}, pages 195--205, 2018.

\bibitem{StickyPillars}
K. Fischer, M. Simon, F. Olsner, S. Milz, H.-M. Gross, and P. Mader.
\newblock {StickyPillars: Robust and Efficient Feature Matching on Point Clouds
  Using Graph Neural Networks}.
\newblock In {\em Conference on Computer Vision and Pattern Recognition
  (CVPR)}, pages 313--323, 2021.

\bibitem{fitzgibbon02RobustReg}
A. Fitzgibbon.
\newblock {Robust Registration of 2D and 3D Point Sets}.
\newblock {\em Image and Vision Computing}, 21:1145--1153, 2002.

\bibitem{KittiOdometry}
A. Geiger, P. Lenz, and R. Urtasun.
\newblock {Are we ready for Autonomous Driving? The KITTI Vision Benchmark
  Suite}.
\newblock In {\em Conference on Computer Vision and Pattern Recognition
  (CVPR)}, 2012.

\bibitem{Gojcic_2020_CVPR}
Z. Gojcic, C. Zhou, J. Wegner, L. Guibas, and T. Birdal.
\newblock {Learning Multiview {3D} Point Cloud Registration}.
\newblock In {\em Conference on Computer Vision and Pattern Recognition
  (CVPR)}, 2020.

\bibitem{Gojcic2019ThePM}
Z. Gojcic, C. Zhou, J. Wegner, and A. Wieser.
\newblock {The Perfect Match: 3D Point Cloud Matching With Smoothed Densities}.
\newblock In {\em Conference on Computer Vision and Pattern Recognition
  (CVPR)}, pages 5540--5549, 2019.

\bibitem{Proscutes}
J. Gower.
\newblock Generalized procrustes analysis.
\newblock {\em Psychometrika}, 40:33--51, 1975.

\bibitem{huang2020predator}
S. Huang, Z. Gojcic, M. Usvyatsov, A. Wieser, and K. Schindler.
\newblock {PREDATOR: Registration of 3D Point Clouds with Low Overlap}.
\newblock In {\em Conference on Computer Vision and Pattern Recognition
  (CVPR)}, pages 4267--4276, 2021.

\bibitem{Izatt17mixedInteger}
G. Izatt, H. Dai, and R. Tedrake.
\newblock {Globally Optimal Object Pose Estimation in Point Clouds with
  Mixed-Integer Programming}.
\newblock In {\em International Symposium on Robotics Research}, 2017.

\bibitem{spin}
A.~E. {Johnson} and M. {Hebert}.
\newblock Using spin images for efficient object recognition in cluttered 3d
  scenes.
\newblock {\em IEEE Transactions on Pattern Analysis and Machine Intelligence},
  21(5):433--449, 1999.

\bibitem{Khoury2017LearningCG}
M. Khoury, Q.-Y. Zhou, and V. Koltun.
\newblock {Learning Compact Geometric Features}.
\newblock In {\em International Conference on Computer Vision (ICCV)}, pages
  153--161, 2017.

\bibitem{lawin2020registration}
F.~J. Lawin and P.-E. Forssén.
\newblock Registration loss learning for deep probabilistic point set
  registration.
\newblock In {\em International Conference on 3D Vision (3DV)}, 2020.

\bibitem{IDAM}
J. Li, C. Zhang, Z. Xu, H. Zhou, and C. Zhang.
\newblock {Iterative Distance-Aware Similarity Matrix Convolution with
  Mutual-Supervised Point Elimination for Efficient Point Cloud Registration}.
\newblock In {\em European Conference on Computer Vision (ECCV)}, 2020.

\bibitem{adamw}
I. Loshchilov and F. Hutter.
\newblock Decoupled weight decay regularization.
\newblock In {\em International Conference on Learning Representations (ICLR)},
  2019.

\bibitem{LucasKanade}
B. Lucas and T. Kanade.
\newblock {An Iterative Image Registration Technique with an Application to
  Stereo Vision}.
\newblock In {\em International joint conference on Artificial intelligence},
  volume~2, pages 674–--679, 1981.

\bibitem{makadia06}
A. Makadia, A. Patterson, and K. Daniilidis.
\newblock Fully automatic registration of 3d point clouds.
\newblock In {\em Conference on Computer Vision and Pattern Recognition
  (CVPR)}, volume~1, pages 1297--1304, 2006.

\bibitem{maron10ConvexRelaxation}
H. Maron, N. Dym, I. Kezurer, S. Kovalsky, and Y. Lipman.
\newblock {Point Registration via Efficient Convex Relaxation}.
\newblock {\em ACM Transactions on Graphics}, 35(4):1--12, 2016.

\bibitem{ReLU}
V. Nair and G.~E. Hinton.
\newblock Rectified linear units improve restricted boltzmann machines.
\newblock In {\em International Conference on Machine Learning (ICML)}, pages
  807--814, 2010.

\bibitem{pomerleau04review}
F. {Pomerleau}, F. {Colas}, and R. {Siegwart}.
\newblock {\em {A Review of Point Cloud Registration Algorithms for Mobile
  Robotics}}.
\newblock 2015.

\bibitem{rosen20}
D.M. Rosen, L. Carlone, A.S. Bandeira, and J.J. Leonard.
\newblock {A Certifiably Correct Algorithm for Synchronization over the Special
  Euclidean Group}.
\newblock In {\em International Workshop on the Algorithmic Foundations of
  Robotics (WAFR)}, volume~13, 2020.

\bibitem{rusinkiewicz01ICPvariants}
S. {Rusinkiewicz} and M. {Levoy}.
\newblock {Efficient variants of the {ICP} algorithm}.
\newblock In {\em International Conference on 3-D Digital Imaging and
  Modeling}, pages 145--152, 2001.

\bibitem{RANSAC}
R.~B. {Rusu}, N. {Blodow}, and M. {Beetz}.
\newblock {Fast Point Feature Histograms (FPFH) for 3D registration}.
\newblock In {\em International Conference on Robotics and Automation (ICRA)},
  pages 3212--3217, 2009.

\bibitem{SHOT}
S. Salti, F. Tombari, and L. Di~Stefano.
\newblock {SHOT: Unique Signatures of Histograms for Surface and Texture
  Description}.
\newblock {\em Computer Vision and Image Understanding}, 125, 2014.

\bibitem{segal09GenICP}
A. Segal, D. Hähnel, and S. Thrun.
\newblock Generalized-{ICP}.
\newblock In {\em Robotics: Science and Systems}, 2009.

\bibitem{silva05}
L. Silva, O.R.P. Bellon, and K.~L. Boyer.
\newblock Precision range image registration using a robust surface
  interpenetration measure and enhanced genetic algorithms.
\newblock {\em IEEE Transactions on Pattern Analysis and Machine Intelligence},
  27(5), 2005.

\bibitem{shape_context}
F. Tombari, S. Salti, and L. Di~Stefano.
\newblock Unique shape context for 3d data description.
\newblock In {\em ACM workshop on 3D object retrieval}, 2010.

\bibitem{tsin2004}
Y. Tsin and T. Kanade.
\newblock A correlation-based approach to robust point set registration.
\newblock In {\em European Conference on Computer Vision (ECCV)}, pages
  558–--569, 2004.

\bibitem{ulyanov2017instance}
D. Ulyanov, A. Vedaldi, and V. Lempitsky.
\newblock Instance normalization: The missing ingredient for fast stylization.
\newblock {\em arXiv:1607.08022}, 2016.

\bibitem{Wang2019PRNetSL}
Y. Wang and J. Solomon.
\newblock {PRNet: Self-Supervised Learning for Partial-to-Partial
  Registration}.
\newblock In {\em Advances in Neural Information Processing Systems (NeurIPS)},
  volume~32, pages 8814--8826, 2019.

\bibitem{Wang2019DeepCP}
Y. Wang and J.~M. Solomon.
\newblock {Deep Closest Point: Learning Representations for Point Cloud
  Registration}.
\newblock In {\em International Conference on Computer Vision (ICCV)}, pages
  3522--3531, 2019.

\bibitem{modelnet40}
Z. Wu, S. Song, A. Khosla, F. Yu, L. Zhang, X. Tang, and J. Xiao.
\newblock {3D ShapeNets: A deep representation for volumetric shapes}.
\newblock In {\em Conference on Computer Vision and Pattern Recognition
  (CVPR)}, pages 1912--1920, 2015.

\bibitem{goicp}
J. {Yang}, H. {Li}, D. {Campbell}, and Y. {Jia}.
\newblock {Go-ICP: A Globally Optimal Solution to 3D ICP Point-Set
  Registration}.
\newblock {\em IEEE Transactions on Pattern Analysis and Machine Intelligence},
  38(11):2241--2254, 2016.

\bibitem{RPM-Net}
Z.~J. Yew and G.~H. Lee.
\newblock Rpm-net: Robust point matching using learned features.
\newblock In {\em Conference on Computer Vision and Pattern Recognition
  (CVPR)}, 2020.

\bibitem{DeepGMR}
W. Yuan, B. Eckart, K. Kim, V. Jampani, D. Fox, and J. Kautz.
\newblock Deepgmr: Learning latent gaussian mixture models for registration.
\newblock In {\em European Conference on Computer Vision (ECCV)}, pages
  733--750, 2020.

\bibitem{zeng20163dmatch}
A. Zeng, S. Song, M. Nie{\ss}ner, M. Fisher, J. Xiao, and T. Funkhouser.
\newblock {3DMatch: Learning Local Geometric Descriptors from RGB-D
  Reconstructions}.
\newblock In {\em Conference on Computer Vision and Pattern Recognition
  (CVPR)}, 2017.

\bibitem{zhang94icp}
Z. Zhang.
\newblock {Iterative point matching for registration of freeform curves and
  surfaces}.
\newblock {\em International Journal on Computer Vision}, 13(2):119--152, 1994.

\bibitem{zhou16FGR}
Q.-Y. Zhou, J. Park, and V. Koltun.
\newblock {Fast Global Registration}.
\newblock In {\em European Conference on Computer Vision (ECCV)}, pages
  766--782, 2016.

\bibitem{zodage20}
T. Zodage, R. Chakwate, V. Sarode, R.~A. Srivatsan, and H. Choset.
\newblock Correspondence matrices are underrated.
\newblock In {\em International Conference on 3D Vision (3DV)}, 2020.

\end{thebibliography}
}

%
\newpage
\appendix
\section{Network Architecture}

\subsection{Multiscale Attention Module $g(\cdot)$}

Fig.~\ref{fig:encoder} shows the detailed architecture of the multiscale attention module and of its $L$ different encoders $e^{(\ell)}: \mathbb{R}^{\cdot \times c^{(\ell - 1)}} \xrightarrow{} \mathbb{R}^{\cdot \times c^{(\ell)}/2}$. We recall that $c^{(\ell)}$ denotes the number of channels at the output of the $\ell^{th}$ multiscale attention module. The encoder $e^{(\ell)}$ is made of three residual blocks, each yielding features with $c^{(\ell)}/2$ channels. The architecture of the residual block is also presented in Fig.~\ref{fig:encoder} where the first 1D convolution is used only when the number of input and output channels differs. Each residual block consists of two FKAConv layers \cite{boulch2020lightconvpoint} with $c^{(\ell)}/2$ channels at input and output, a neighborhood of $32$ nearest neighbours, a stride of size $1$ (no downsampling of the point clouds), and a kernel of size $16$. All convolutional layers are followed by instance normalization (IN) \cite{ulyanov2017instance} with learned affine correction and a ReLU activation \cite{ReLU}. The number of channels $c^{(\ell)}$ used in our experiments in given in Table~\ref{tab:cell}.

\begin{table}[h]
\begin{center}
\ra{1.1}
\small
\setlength{\tabcolsep}{5pt}
\begin{tabular}{@{} l r r r r r r r @{}}
\toprule
$\ell$ & 
    $0$ & 
    $1$ & 
    $2$ & 
    $3$ & 
    $4$ & 
    $5$ & 
    $6$
\\
\midrule
$c^{(l)}$ & 
    $3$  & 
    $32$ & 
    $32$ & 
    $64$ & 
    $64$ &
    $128$ &
    $128$
\\
\bottomrule
\end{tabular}
\end{center}
\caption{\textbf{Number of channels $c^{(\ell)}$ at the ouput of the $\ell^\th$ block in the multiscale attention module}. $c^{(0)} = 3$ corresponds to the three coordinates $x, y, z$ of the input point cloud. }
\label{tab:cell}
\end{table}
%

\subsection{Confidence Estimator $h(\cdot, \cdot)$}

\begin{figure*}[t]
    \begin{center}
    \includegraphics[width=\linewidth]{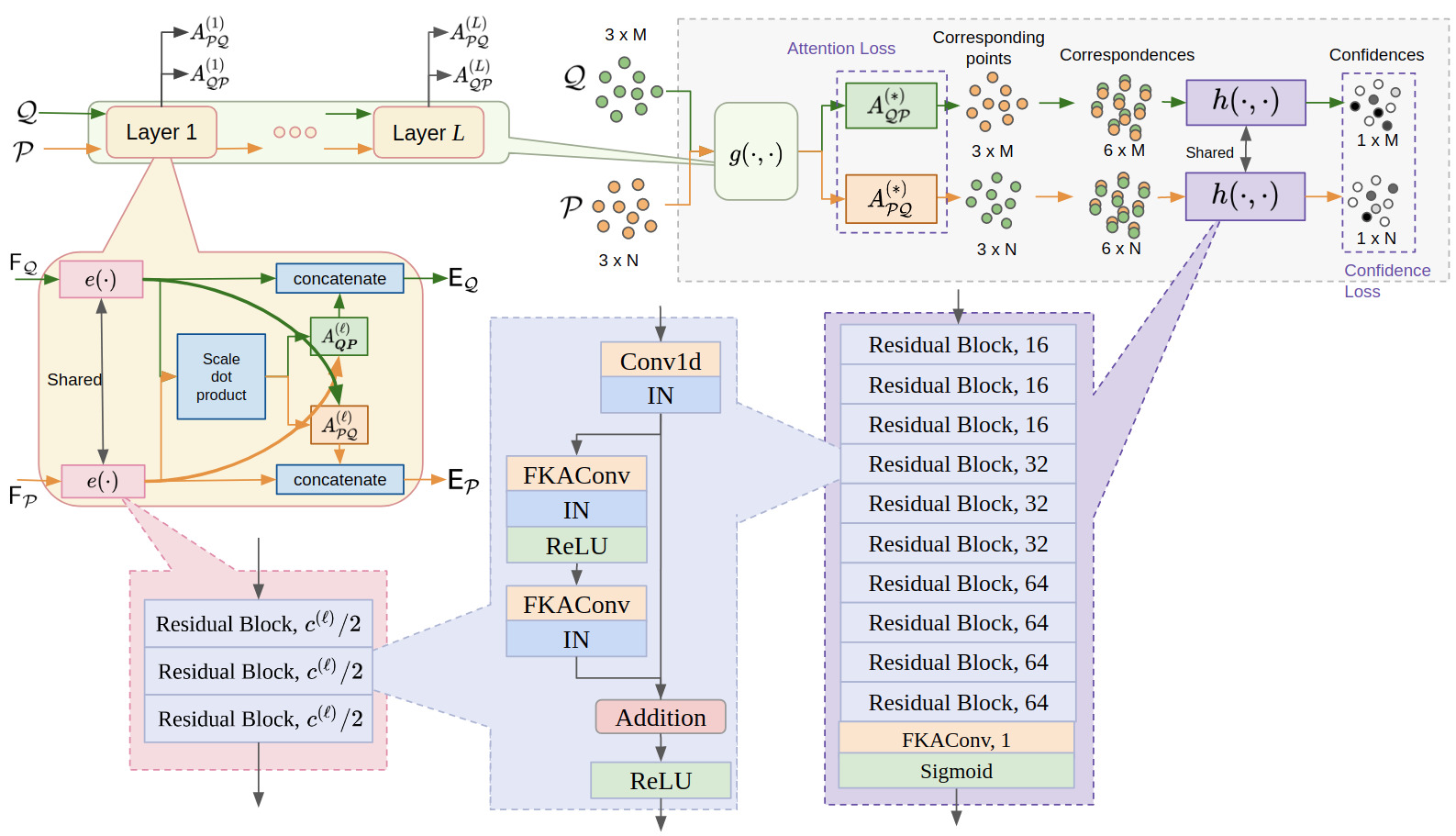}
    \caption{\textbf{Detailed network architecture of the multiscale attention module $g(\cdot)$ and the confidence estimator module $h(\cdot, \cdot)$}. Each module is constructed using the same type residual block, in which all internal layers have the same number of channels $C$ indicated in the notation ``Residual block, $C$''. Note that the 1D convolution in the residual block is used \emph{only} when the number of input and output channels differ.}
    \label{fig:encoder}
    \end{center}
\end{figure*}

Fig.~\ref{fig:encoder} shows the detailed architecture of the confidence estimator $h: \mathbb{R}^{\cdot \times 6} \xrightarrow[]{} (0, 1)^{\cdot \times 1}$. It consists of nine residual blocks, a FKAConv layer that reduces the number of channels to $1$ and a final sigmoid activation. The FKAConv layer has a neighborhood of 32 nearest neighbours, a stride of size $1$, and a kernel of size $16$. The residual blocks have the same structure as the one used to construct the encoders in the multiscale attention module.

%

%
\section{Experimental results}

\subsection{Indoor dataset: 3DMatch}

\begin{figure*}
    \begin{center}
    \includegraphics[width=\linewidth]{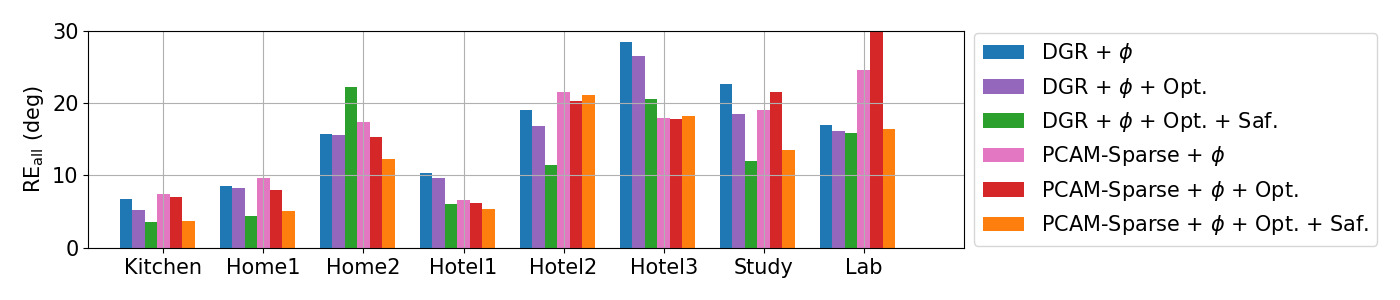}
    \includegraphics[width=\linewidth]{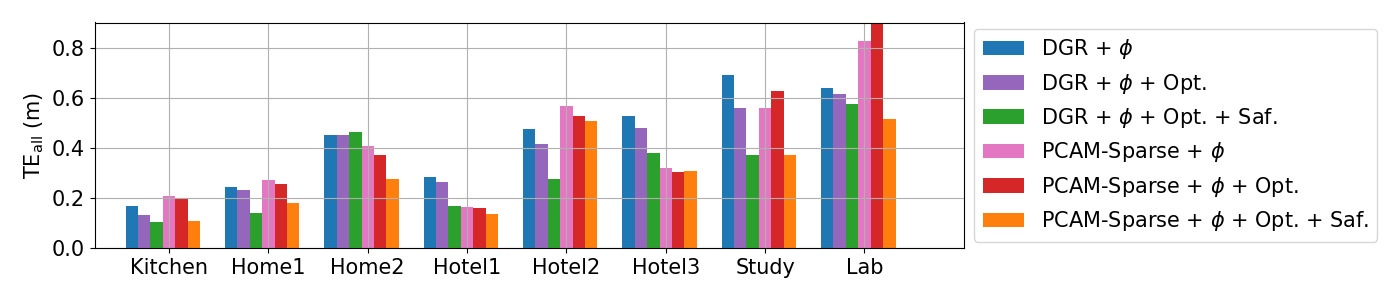}
    \includegraphics[width=\linewidth]{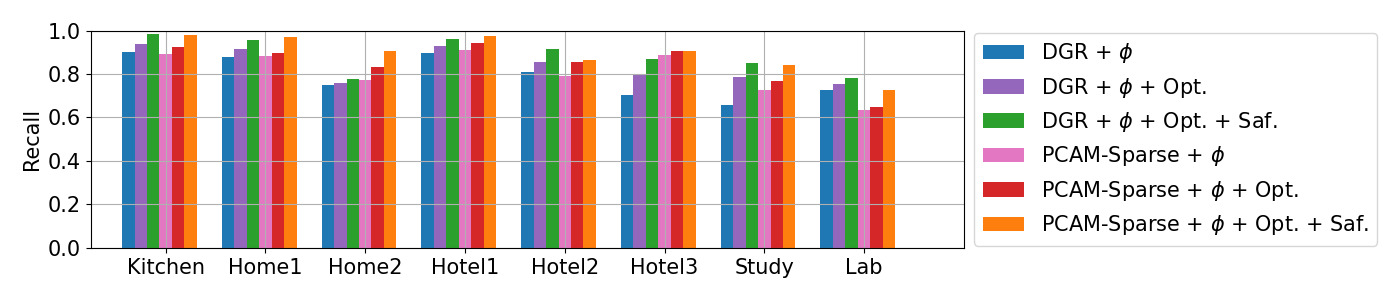}
    \includegraphics[width=\linewidth]{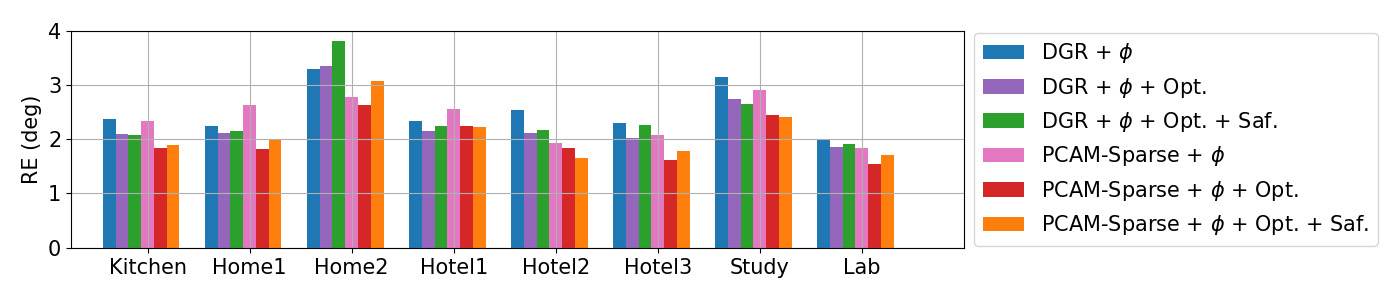}
    \includegraphics[width=\linewidth]{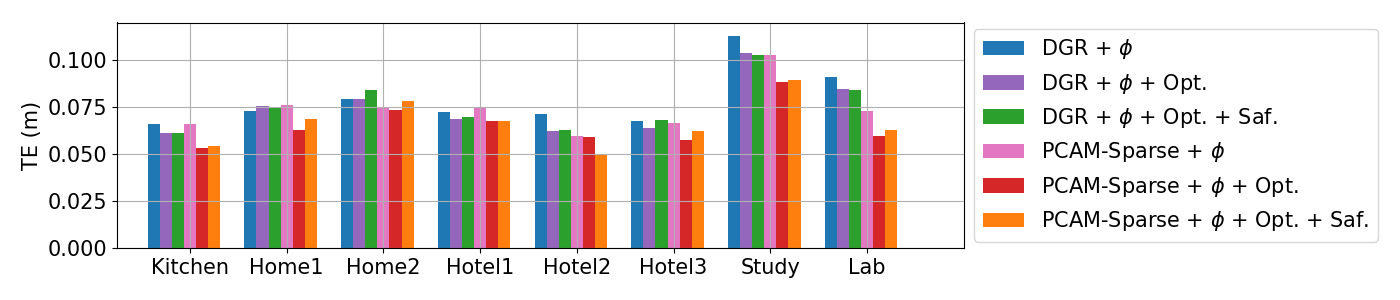}
    \caption{\textbf{Analysis of 3DMatch registration results per scene}. Row 1-2: Average TE and RE measured \emph{on all pairs} (lower is better). Row~3: recall rate (higher is better). Row 4-5: TE and RE measured \emph{on successfully registered pairs} (lower is better). Note that the recall of two methods have to be almost identical for their errors RE (resp.\ TE) to be comparable.}
    \label{fig:recall_per_scene_3dmatch}
    \end{center}
\end{figure*}

A detailed analysis of the performance of our method for each of the $8$ scenes in the test set of 3DMatch \cite{zeng20163dmatch} is presented in Fig.~\ref{fig:recall_per_scene_3dmatch}. The results confirm the global scores presented in the main part of the paper. Each of the post-processing steps permits to improve the recall on all the scenes. DGR and PCAM-Sparse achieve nearly similar recall on all scenes with a very noticeable difference on Hotel3 where PCAM-Sparse with hard-thresholding on the weights outperforms DGR with all post-processing steps used.

\begin{figure*}
    \begin{center}
    \includegraphics[width=.99\linewidth]{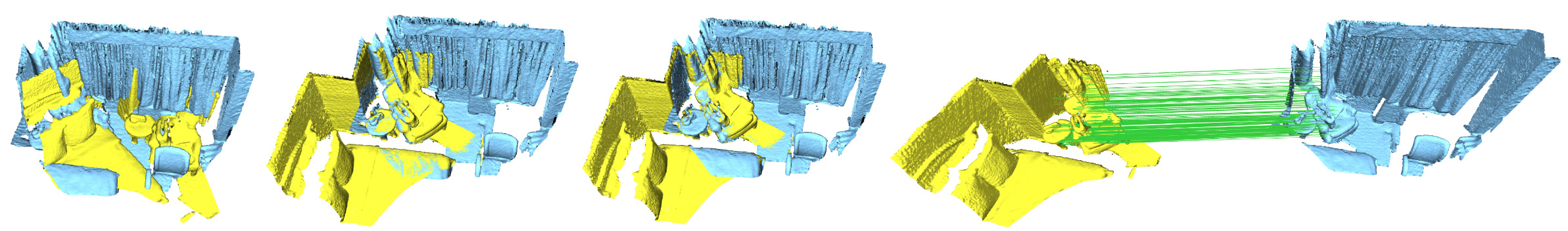}
    \includegraphics[width=.99\linewidth]{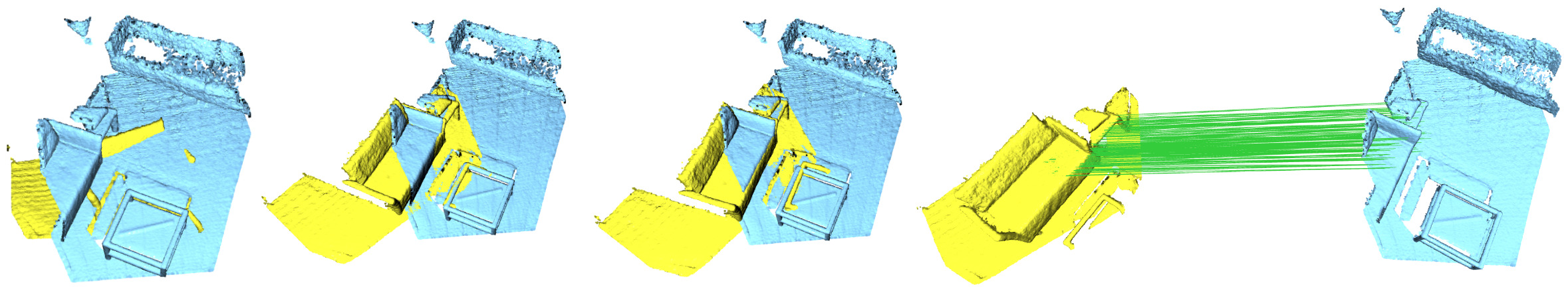}
    \includegraphics[width=.99\linewidth]{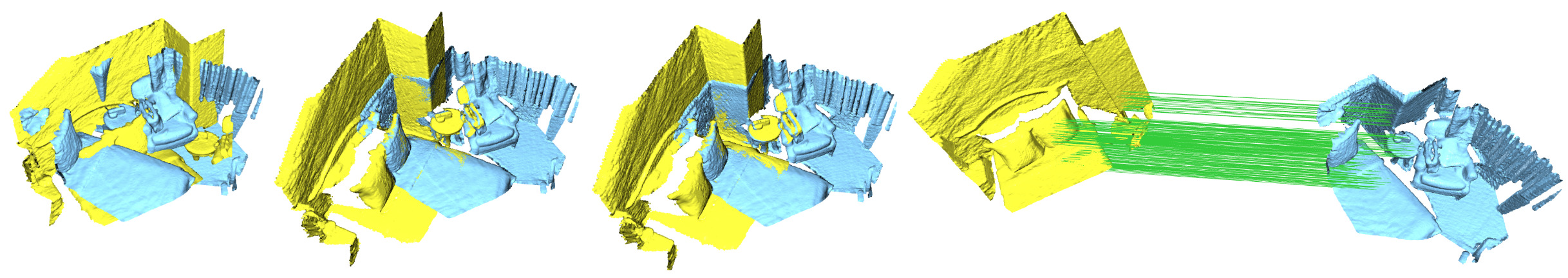}
    \includegraphics[width=.99\linewidth]{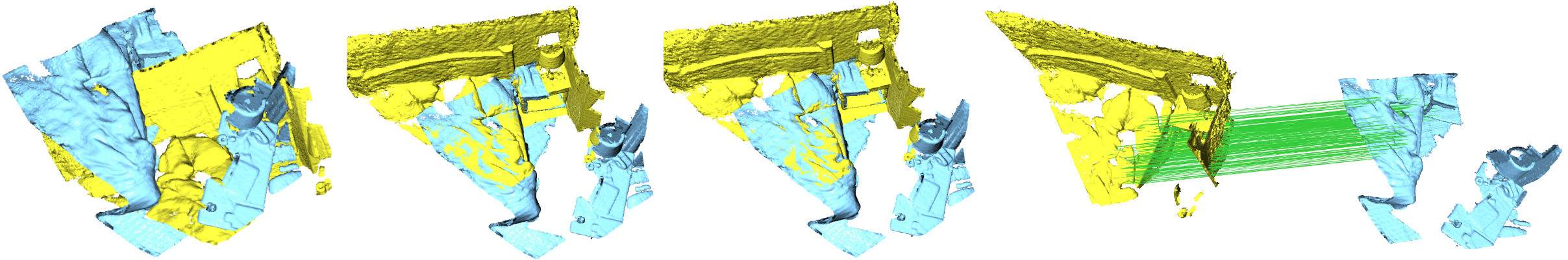}
    \includegraphics[width=.99\linewidth]{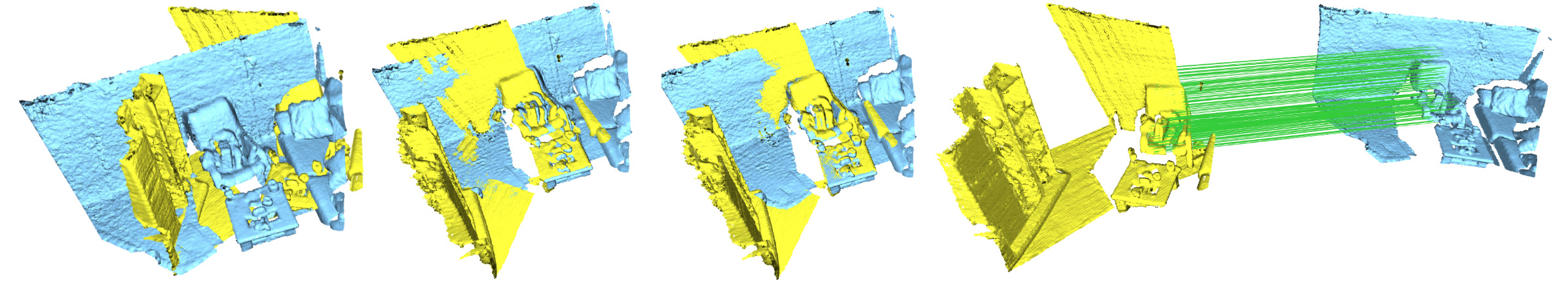}
    \includegraphics[width=.99\linewidth]{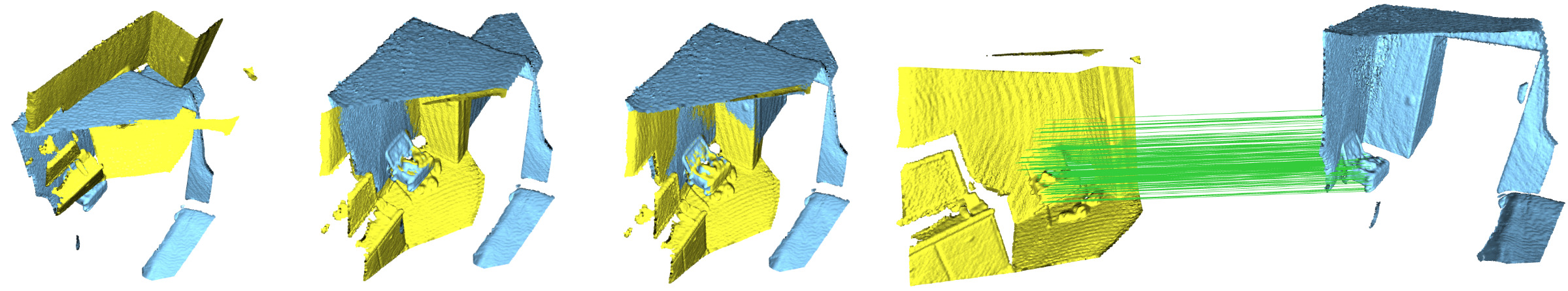}
    \includegraphics[width=.99\linewidth]{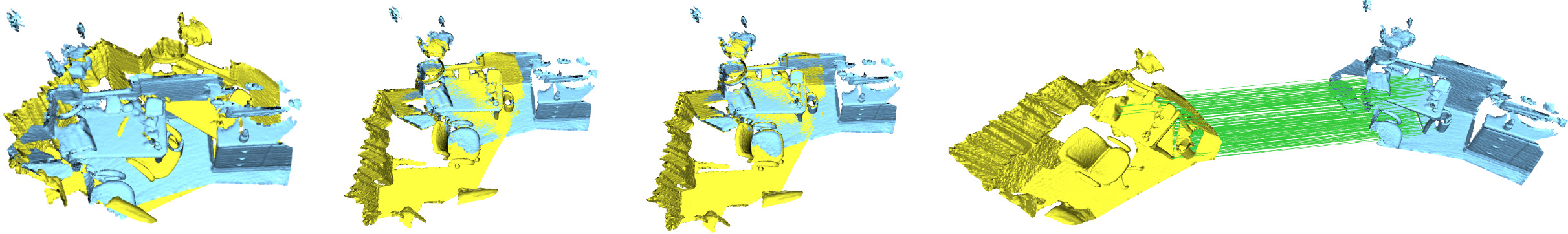}
    \caption{\textbf{Examples of successful registrations on 3DMatch with PCAM-Sparse + $\phi$}. From left to right: overlaid, non-registered input scans (blue and yellow colors); ground-truth registration; scans registered with our method; and top 256 pairs of matched points with highest confidence.}
    \label{fig:3dmath_success_registrations}
    \end{center}
\end{figure*}

\begin{figure*}
    \begin{center}
    \includegraphics[width=\linewidth]{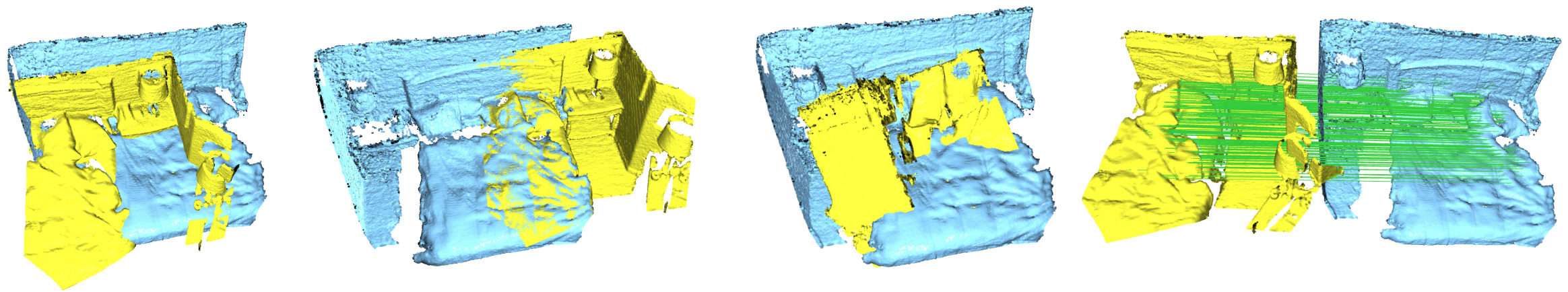}
    \includegraphics[width=\linewidth]{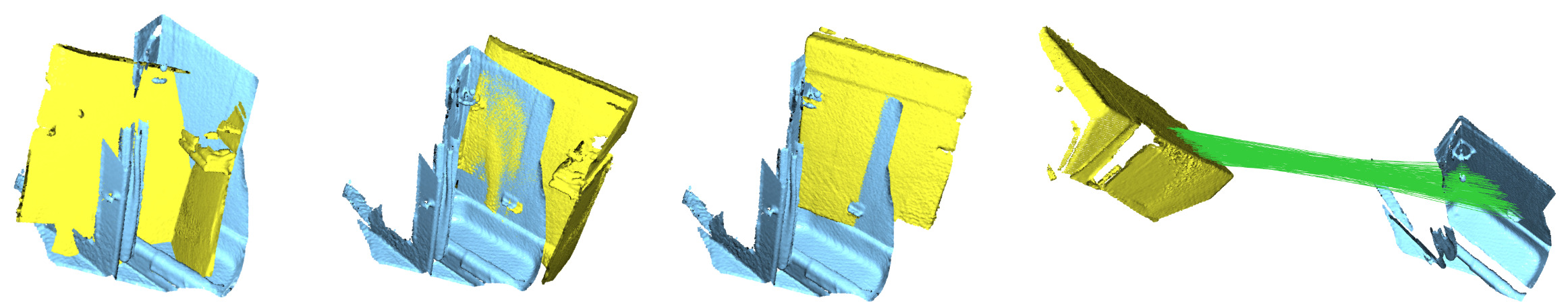}
    \includegraphics[width=\linewidth]{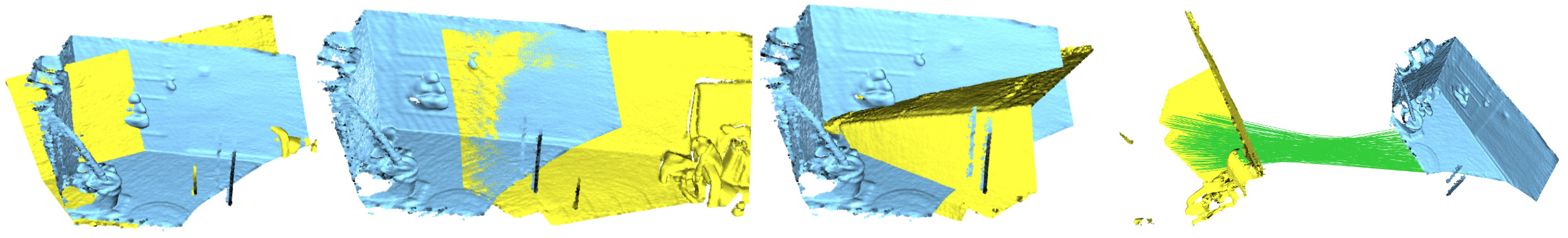}
    \includegraphics[width=\linewidth]{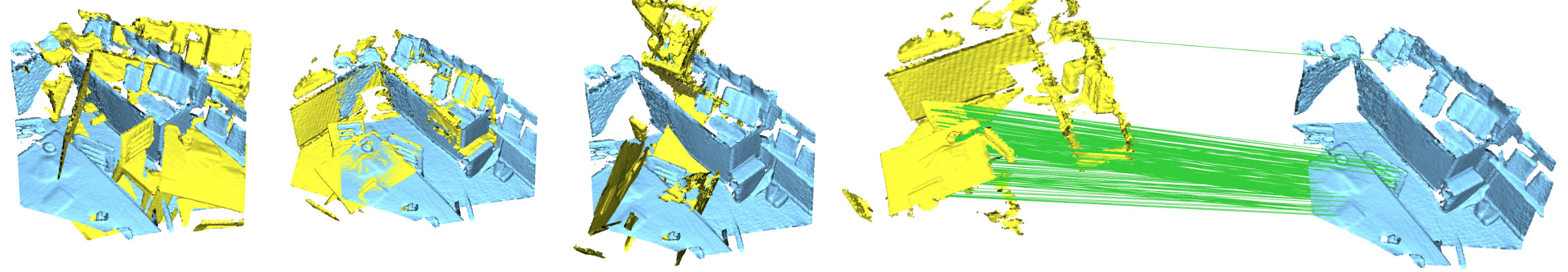}
    \includegraphics[width=\linewidth]{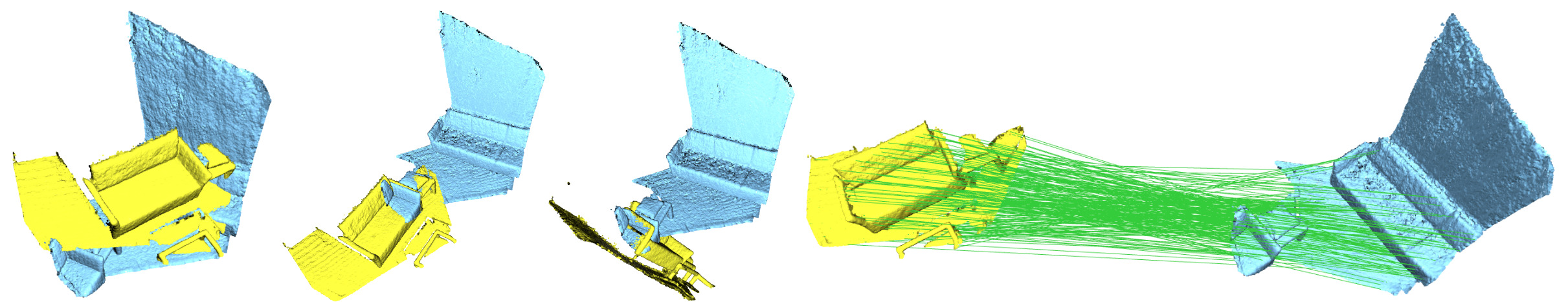}
    \includegraphics[width=\linewidth]{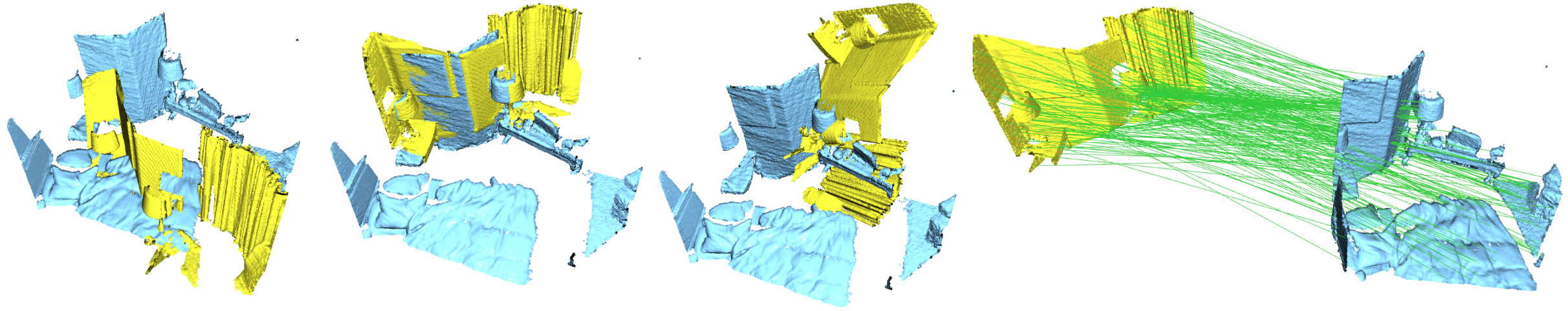}
    \caption{\textbf{Examples of failed registrations on 3DMatch with PCAM-Sparse + $\phi$}. From left to right: overlaid, non-registered input scans (blue and yellow colors); ground-truth registration; scans registered with our method; and top 256 pairs of matched points with highest confidence.}
    \label{fig:3dmatch_failed_registrations}
    \end{center}
\end{figure*}

We present in Fig.~\ref{fig:3dmath_success_registrations} and Fig.~\ref{fig:3dmatch_failed_registrations} examples of success and failure, respectively, of PCAM-Sparse with filtering with $\phi$ on 3DMatch \cite{zeng20163dmatch}. We notice that the misregistrations are due to wrong mappings between points on planar surfaces in the examples of row $2$, $3$ in Fig.~\ref{fig:3dmatch_failed_registrations}. We also remark in the example at row $5$ that our method found matching points between points of similar but different object. Finally, row $1$ of Fig.~\ref{fig:3dmatch_failed_registrations} shows a case where the estimated correspondences are correct but the confidence estimator considers them as unreliable which leads to a wrong estimation of the transformation (unless the safeguard is activated). This shows that the performance of the confidence estimator can still be improved.

\begin{table*}[t]
\begin{center}
\ra{1.1}
\setlength{\tabcolsep}{2.5pt}
\footnotesize
\begin{tabular}{@{} l r r r r r r r r r r r r @{}}
\toprule
 & \multicolumn{4}{c}{Unseen objects}
 & \multicolumn{4}{c}{Unseen categories}
 & \multicolumn{4}{c}{Unseen objects + noise}
\\
\cmidrule(lr){2-5}
\cmidrule(lr){6-9}
\cmidrule(lr){10-13}
Method & 
    RMSE (\ma{R}) & MAE (\ma{R}) & 
    RMSE (\vec{t}) & MAE (\vec{t}) &
    RMSE (\ma{R}) & MAE (\ma{R}) & 
    RMSE (\vec{t}) & MAE (\vec{t}) &
    RMSE (\ma{R}) & MAE (\ma{R}) & 
    RMSE (\vec{t}) & MAE (\vec{t})
\\
\midrule
PointNetLK \cite{PointNetLK} &
    16.73 & 7.55 &
    0.045 & 0.025 &
    22.94 & 9.65 &
    0.061 & 0.033 &
    19.94 & 9.08 & 
    0.057 & 0.032
\\
DCP-v2 \cite{Wang2019DeepCP} & 
    6.71 & 4.45 &
    0.027 & 0.020 &
    9.77 & 6.95 &
    0.034 & 0.025 &
    6.88 & 4.53 & 
    0.028 & 0.021
\\
PRNet \cite{Wang2019PRNetSL} & 
    3.20 & 1.45 &
    0.016 & 0.010 &
    4.98 & 2.33 &
    0.021 & 0.015 &
    4.32 & 2.05 & 
    0.017 & 0.012
\\
IDAM \cite{IDAM} &
    2.46 & 0.56	&
    0.016 &  0.003 &
    3.04 & 0.61	&
    0.019 &  0.004 & 
    3.72 & 1.85 & 
    0.023 & 0.011
\\
OPRNet \cite{3D-3D} &
    0.328	& 0.052 &	
    0.002 & 0.0003 &
    0.357	& 0.069 &	
    0.002 & 0.0004 &
    2.06 & 0.68 & 
    0.008 & 0.003
\\
PCAM-Sparse &
    \bf 0.023 & 0.012 &
    0.0002 & 0.00009 &
    0.072 & 0.018 &
    \bf 0.0002 & \bf 0.0001 &
    0.58 & \bf 0.15 &
    \bf 0.002 & \bf 0.001
\\
PCAM-Soft &
    \bf 0.023 & \bf 0.006 &
    \bf 0.0001 & \bf 0.00004 & 
    \bf 0.056 & \bf 0.013 &
    0.0006 & \bf 0.0001 &
    \bf 0.47 & \bf 0.15 &
    \bf 0.002 & \bf 0.001
\\
\bottomrule
\end{tabular}
\end{center}
\caption{\textbf{Results on ModelNet40 for unseen objects, unseen categories and unseen objects with Gaussian noise.} The scores are reported from \cite{Wang2019PRNetSL} for the first 3  methods, and from the associated papers for the others.}
\label{tab:modelnet_unseen_objects}
\end{table*}
%

\subsection{Outdoor dataset: KITTI}

\begin{figure*}
    \begin{center}
    \includegraphics[width=.99\linewidth]{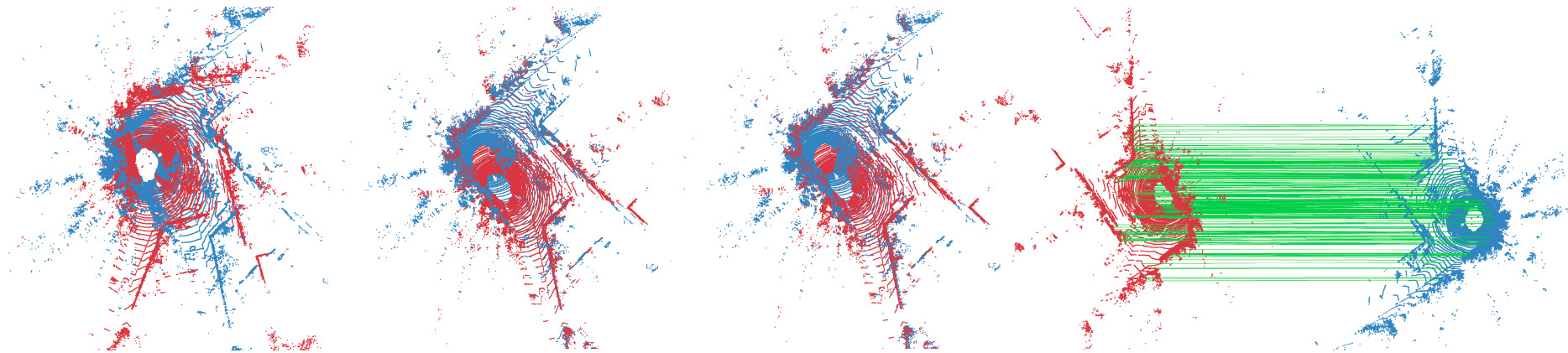}\\
    \vspace{3mm}
    \includegraphics[width=.99\linewidth]{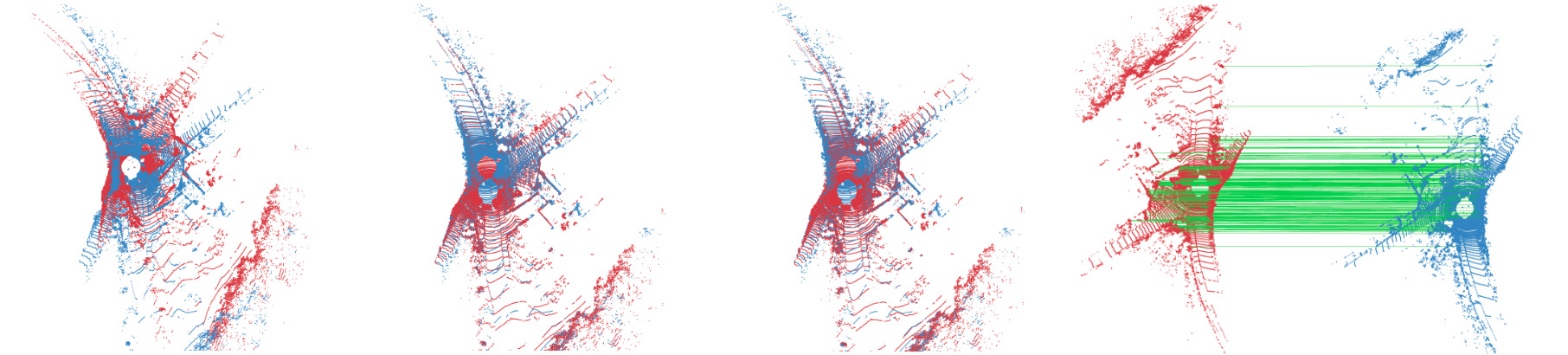}\\
    \vspace{3mm}
    \includegraphics[width=.99\linewidth]{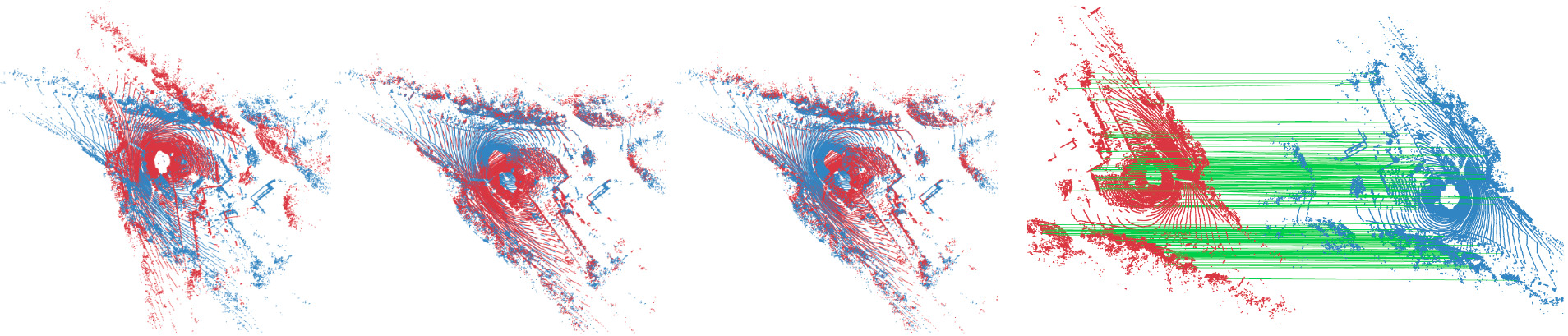}\\
    \vspace{3mm}
    \includegraphics[width=.99\linewidth]{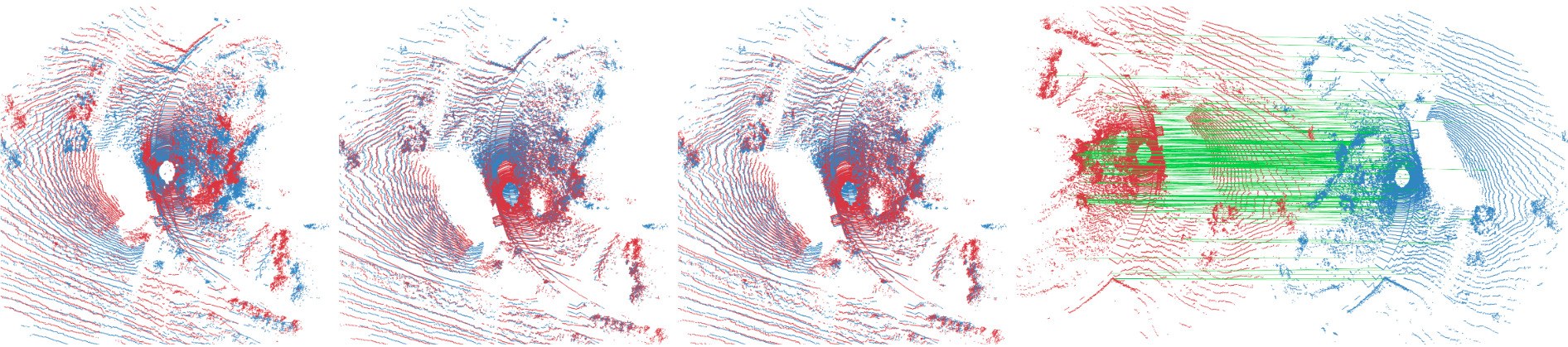}\\
    \vspace{3mm}
    \vspace{3mm}
    \includegraphics[width=.99\linewidth]{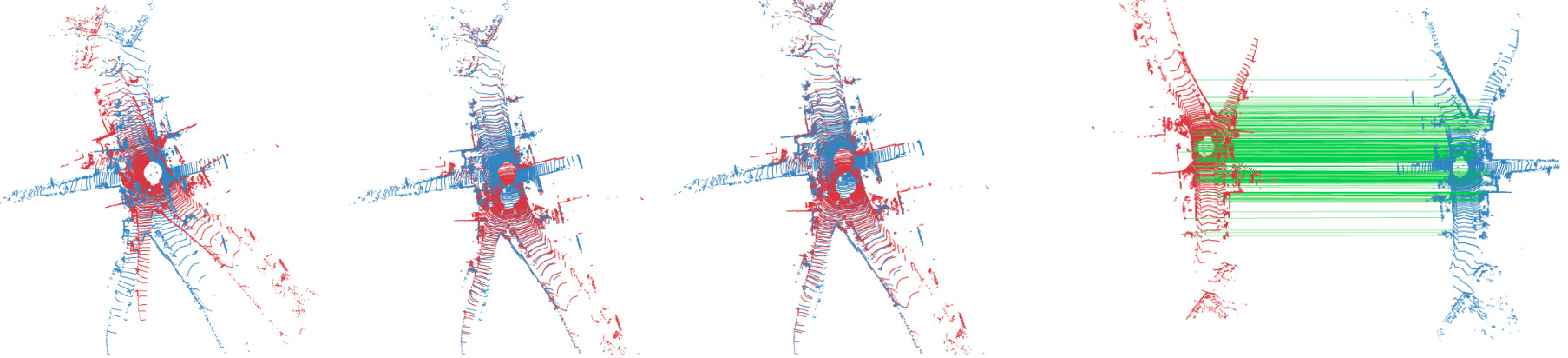}
    \caption{\textbf{Examples in bird-eye view of successful registrations on the KITTI odometry dataset \cite{KittiOdometry}}. From left to right: overlaid, non-registered input  scans (blue and red colors); ground-truth registration; scans registered with our method; and top 256 pairs of matched points with highest confidence. \emph{Note that we have used the full non-voxelized Lidar scans for a better visualisation. However, all registrations are done using $2048$ points drawn at random after voxelization of the full point cloud}.}
    \label{fig:kitti_registrations}
    \end{center}
\end{figure*}

\begin{figure*}
    \begin{center}
    \includegraphics[width=.98\linewidth]{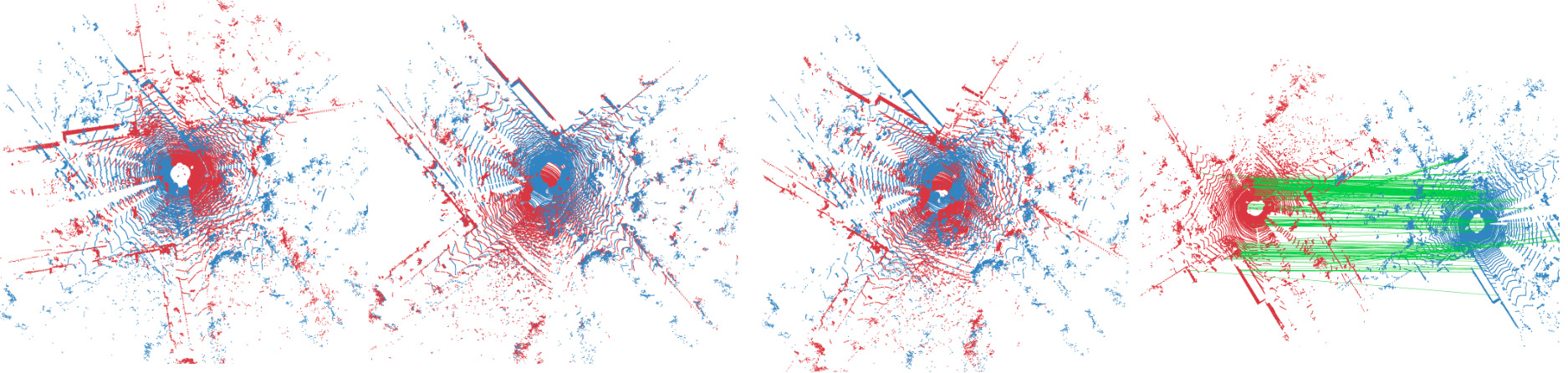}\\
    \vspace{1mm}
    \includegraphics[width=.98\linewidth]{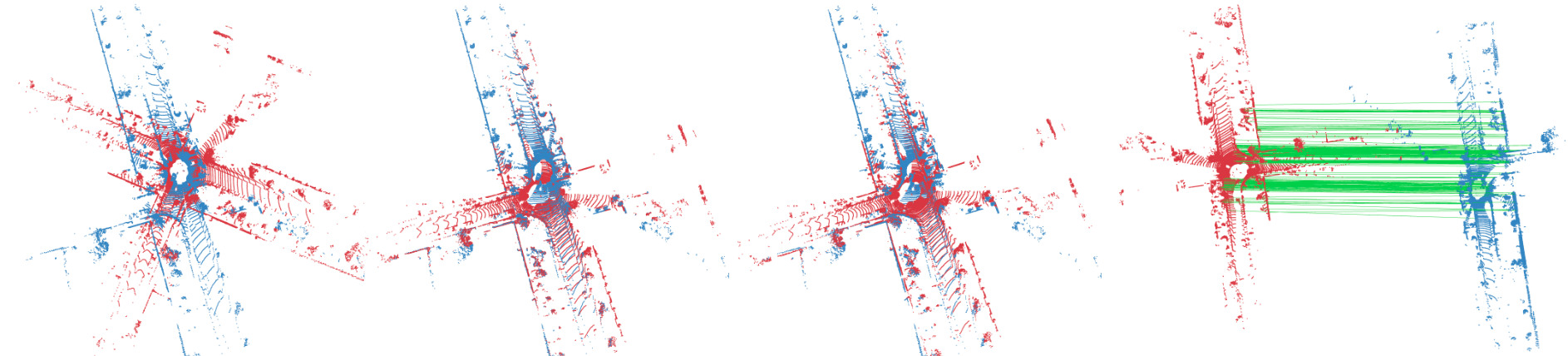}\\
    \vspace{1mm}
    \includegraphics[width=.98\linewidth]{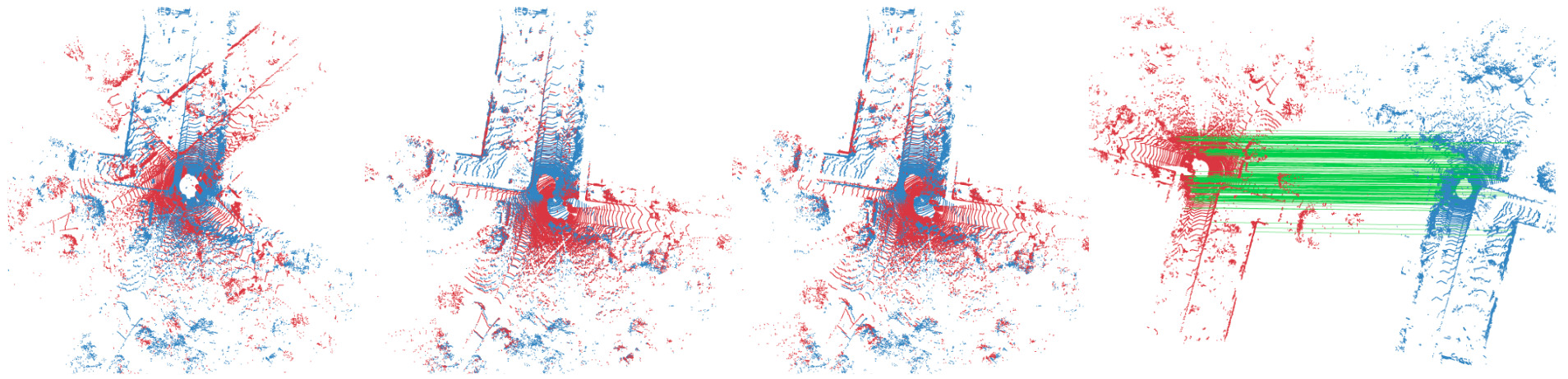}\\
    \vspace{1mm}
    \includegraphics[width=.98\linewidth]{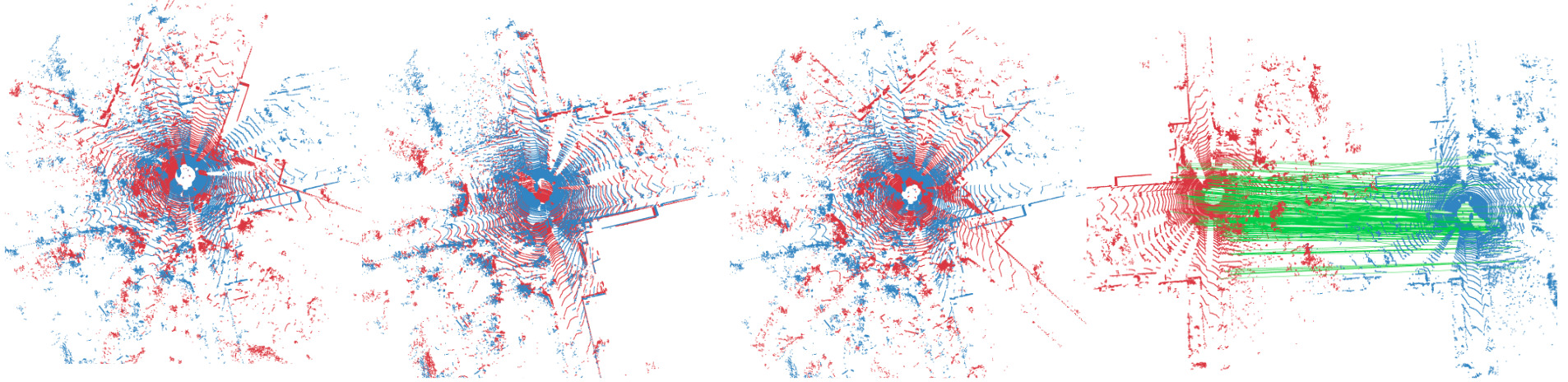}\\
    \vspace{1mm}
    \includegraphics[width=.98\linewidth]{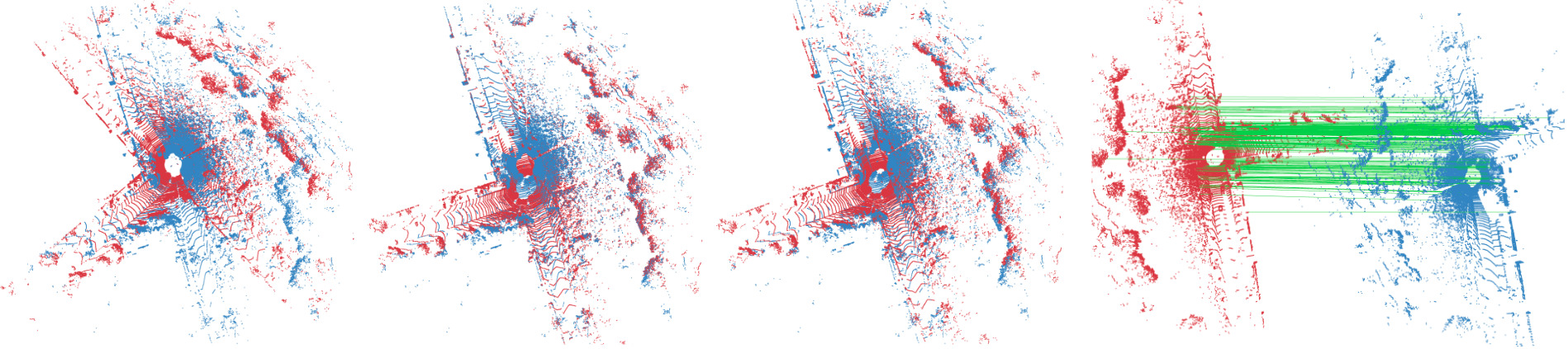}\\
    \caption{\textbf{Examples in bird-eye view of failed registrations on the KITTI odometry dataset \cite{KittiOdometry}}. From left to right: overlaid, non-registered input scans (blue and red colors); ground-truth registration; scans registered with our method; and top 256 pairs of matched points with highest confidence. \emph{Note that we have used the full non-voxelized Lidar scans for a better visualisation. However, all registrations are done using $2048$ points drawn at random after voxelization of the full point cloud}.}
    \label{fig:kitti_failed_registrations}
    \end{center}
\end{figure*}

We present in Fig.~\ref{fig:kitti_registrations} and Fig.~\ref{fig:kitti_failed_registrations} examples of successful and failed registration results, respectively, with our method on the KITTI odometry dataset \cite{KittiOdometry}. We observe in Fig.~\ref{fig:kitti_registrations} that the static objects are well aligned after registrations. Concerning the failed registration results, it seems that this is due to mapping between similar but different structure in the scene, at least for the scans with very large registration errors such as on row $1$ or $4$.

%

\subsection{Synthetic dataset: ModelNet40}

We provide in Table~\ref{tab:modelnet_unseen_objects} the results obtained for ModelNet40 on the split corresponding to unseen objects, unseen categories, and unseen objects with noise \cite{Wang2019PRNetSL}. Our method achieves better results than the recent methods that provided scores on these versions of the dataset.

RPM-Net \cite{RPM-Net} is also evaluated on ModelNet40, but using a slightly different setting than the one used by the methods in Table~\ref{tab:modelnet_unseen_objects}. We also test PCAM on this variant of ModelNet40. RPM-Net uses several passes/iterations to align two point clouds while PCAM, which is not trained on small displacements for refinement, uses only one. For fairness, we evaluate both methods after the first main pass. PCAM outperforms RPM-Net on the ‘clean’ version of ModelNet40 (Chamfer error of $1.8\cdot10^{-5}$ for RPM-Net, $3.4\cdot10^{-9}$ for PCAM) and on its `noisy' version ($7.9\cdot10^{-4}$ for RPM-Net, $6.9\cdot10^{-4}$ for PCAM).

\begin{figure*}
    \begin{center}
    \includegraphics[width=\linewidth]{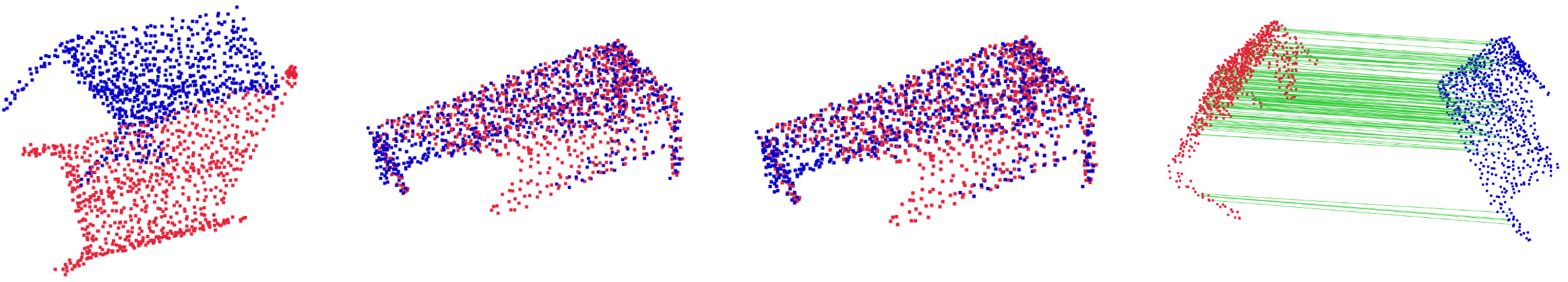}
    \includegraphics[width=\linewidth]{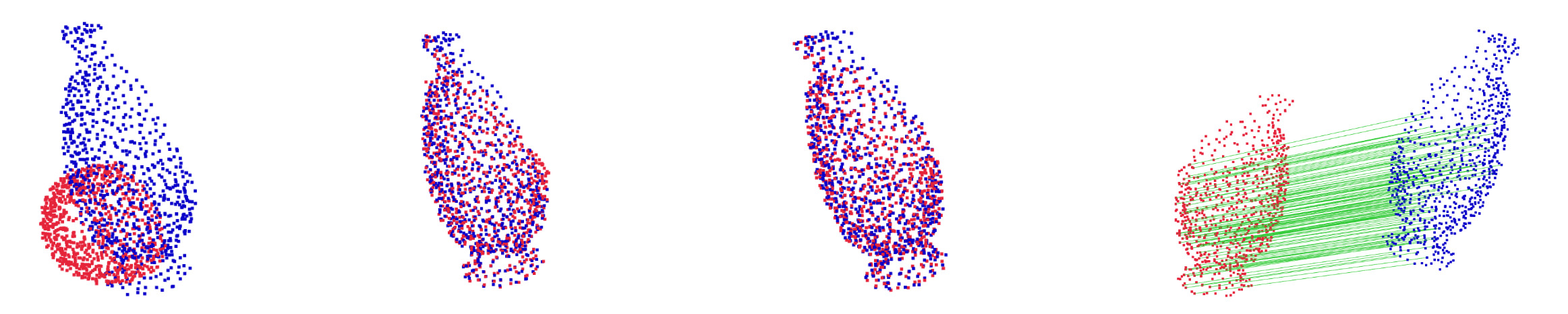}
    \includegraphics[width=\linewidth]{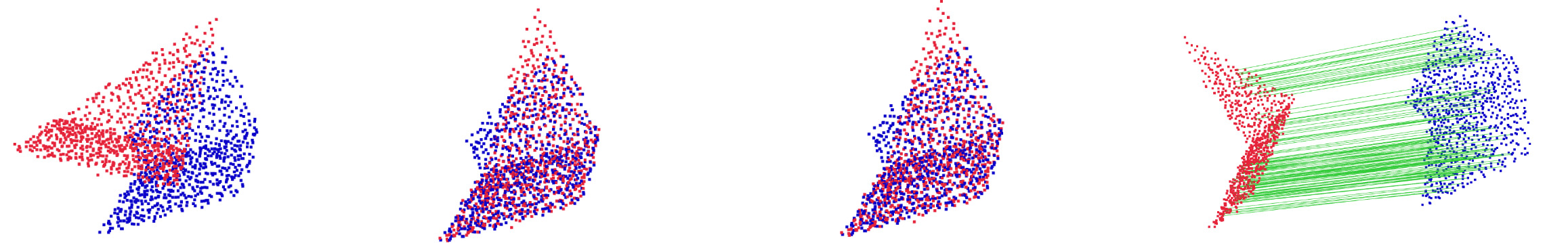}
    \includegraphics[width=\linewidth]{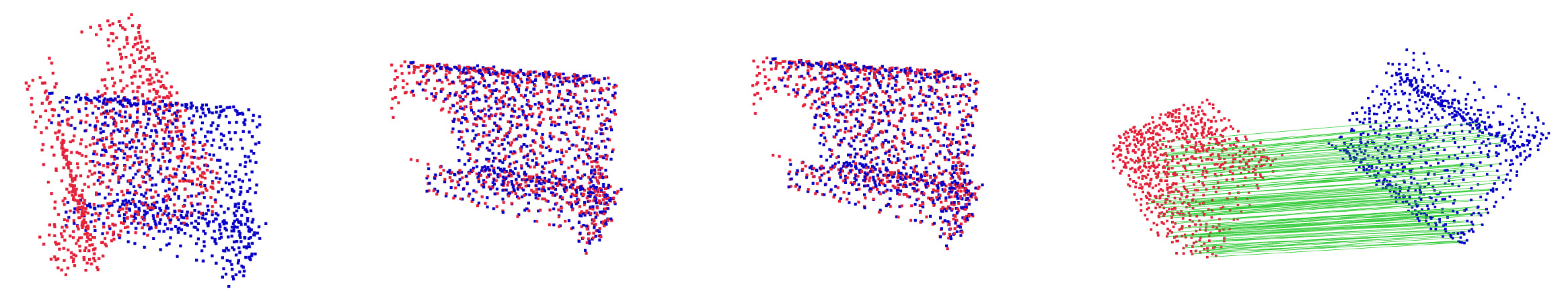}
    \includegraphics[width=\linewidth]{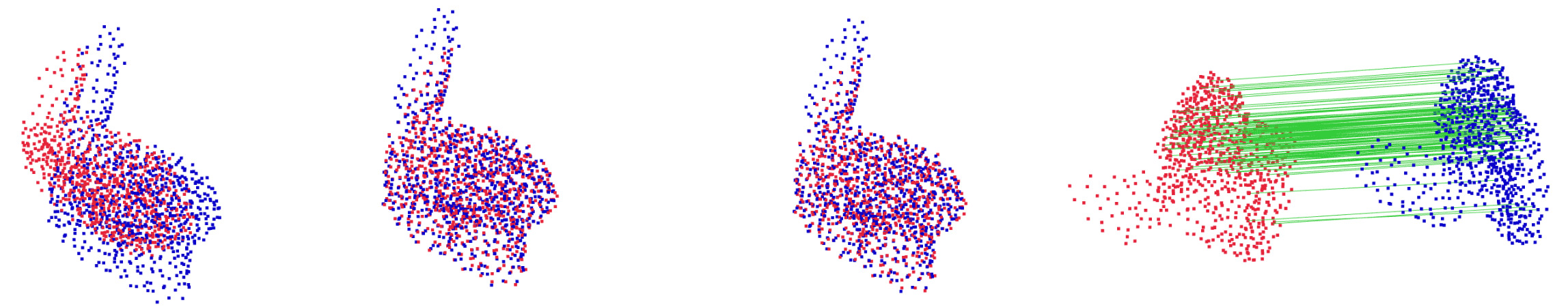}
    \includegraphics[width=\linewidth]{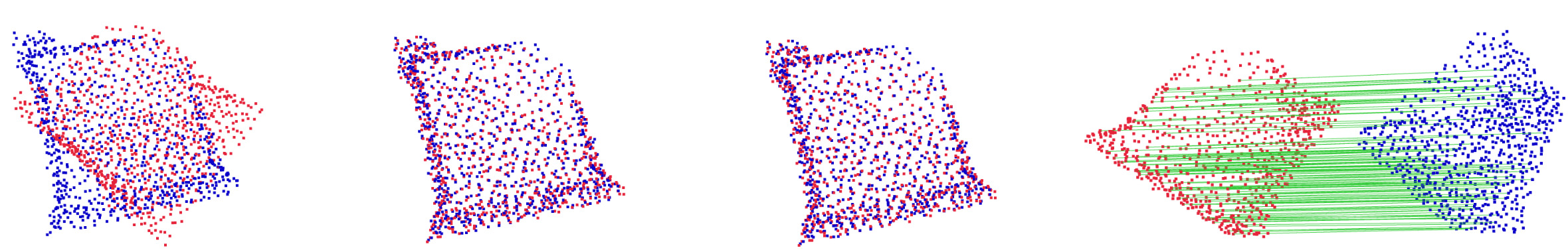}
    \caption{\textbf{Examples of registration results on ModelNet40 on the split with unseen objects, with Gaussian noise \cite{Wang2019PRNetSL}}. From left to right: overlaid, non-registered input scans (blue and red colors); ground-truth registration; scans registered with our method; and top 128 pairs of matched points with highest confidence.}
    \label{fig:modelnet_object_noise}
    \end{center}
\end{figure*}

We present in Fig.~\ref{fig:modelnet_object_noise} examples of registration results with our method for pairs of scans in the unseen objects with Gaussian noise split. The results presented in these figures illustrate the accurate registrations as well as the good quality of the matched pairs of points.

\subsection{Sparse vs FKAConv convolutions}

DGR uses sparse convolutions, whereas PCAM uses FKAConv point convolutions. To test if DGR suffers from a large disadvantage due to these sparse convolutions, we experiment replacing our point matching network by DGR's pre-trained point matching network. We then retrain our confidence estimator on KITTI using DGR's matched points. The performance of this new system reaches a recall of $94.6\%$, an $\rm RE_{all}$ of $2.9$ and $\rm TE_{all}$ of $0.46$ on KITTI validation set. This is on par with the results obtained with our point matching network, showing using sparse convolutions in the point matching network do not perform much worse than FKAConv.

\end{document}